\documentclass[journal]{IEEEtran}

\usepackage{amsmath,amssymb}
\usepackage{graphicx}
\usepackage{booktabs}
\usepackage{multirow}
\usepackage{array}
\usepackage{url}
\usepackage[hidelinks]{hyperref}
\usepackage{xcolor}

\begin{document}

\title{When Do Language-Grounded Explanations Help? A Graph-Bottleneck for Farm Monitoring Interpretable Sheep Facial Pain}
\author{\IEEEauthorblockN{Alam Noor\IEEEauthorrefmark{1} and
Miguel Gutiérrez Gaitán\IEEEauthorrefmark{5} 
}

\IEEEauthorblockA{\IEEEauthorrefmark{1}CISTER Research Center, Porto, Portugal.}
\IEEEauthorblockA{\IEEEauthorrefmark{5}Department of Electrical Engineering, Pontificia Universidad Católica de Chile, Santiago 7820436, Chile.}


\thanks{\textit{(Corresponding author: Alam Noor)}}}


\maketitle

\begin{abstract}
Automated pain recognition from facial expression could make continuous welfare assessment practical in sheep, but adoption depends on trust: a stockperson cannot act on a score that arrives without justification. We ground a model in the Sheep Pain Facial Expression Scale (SPFES) by letting each detected facial region attend over text embeddings of the clinical descriptors and then test whether the resulting explanations mean anything. They do not. Ablating an entire descriptor changes the predicted logit by about $10^{-4}$, and the most-attended cue agrees with the predicted pain level in only $32.6\%$ of regions, although the attention maps, the learned gate, and the generated text all proposed otherwise. We therefore remove the appearance bypass with a concept bottleneck whose classifier reads only SPFES concept scores, supervised by per-region state annotations that image-level pipelines discard. This costs $0.05$--$0.10$ in Cohen's $\kappa$ but yields concepts that are demonstrably learned: minority pain-indicating states are recovered at $3.5$--$8.3\times$ their base rates, and the ear and eye severity orderings emerge without severity supervision. Removing the supervision alone leaves $\kappa$ unchanged while concept accuracy falls to $0.109$, showing that architectural necessity does not imply semantic validity. We also show that pooled concept accuracy is misleading under clinical imbalance and provide a cross-validated, protocol-matched benchmark of seven methods on this dataset.
\end{abstract}



\begin{figure}[!t]
\centering
\includegraphics[width=\linewidth]{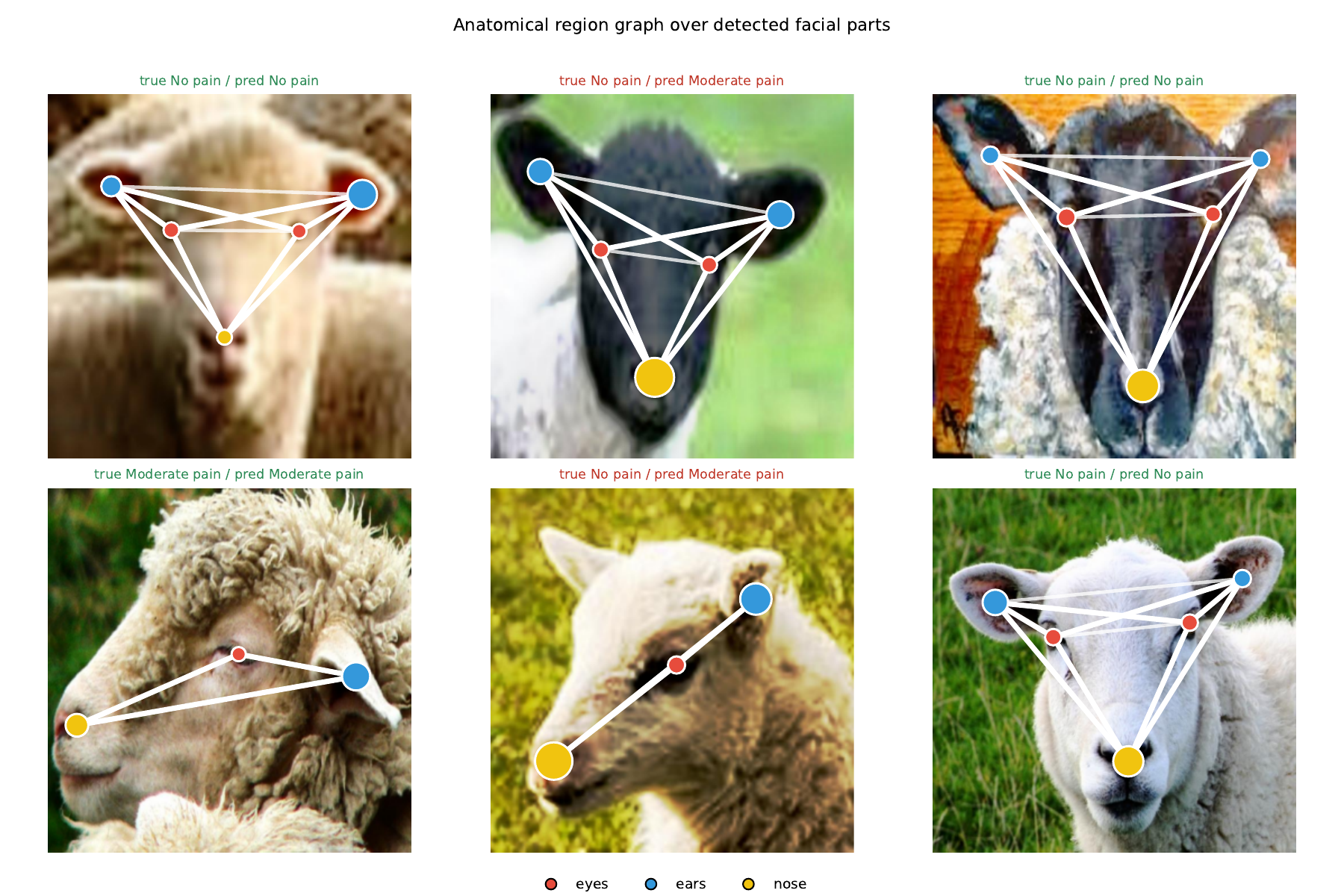}
\caption{The anatomical region graph drawn over detected facial parts. Nodes are region centroids coloured by type (eyes, ears, nose); edge width is
proportional to the anatomical weight $A_{ij}$ used by the normalized filter of \eqref{eq:filter}. The graphs are small and densely connected, so a single message-passing hop already gives every region access to the whole face.}
\label{fig:overlay}
\end{figure}
\section{Introduction}
\label{sec:intro}

\IEEEPARstart{P}{ain} in sheep is prevalent, severe, and often inadequately treated. Conditions like footrot and mastitis are prevalent in commercial flocks, and due to sheep being a prey species that hides visible symptoms of discomfort, those animals demanding the greatest assistance are sometimes the most challenging to detect by casual observation. The Sheep Pain Facial Expression Scale (SPFES) \cite{MCLENNAN201619} addressed this issue by categorizing pain into a narrow spectrum of designated facial changes; orbital constriction, ear positioning, and muzzle tension that experienced evaluators can assess reliably. The system works, although it lacks scalability: evaluation necessitates a skilled evaluator and careful monitoring of each area of sheep faces, which is at odds with the group sizes and labor limitations of commercial production \cite{NOOR2023100366}.

The automation of SPFES scoring presents an attractive research goal, and a number of studies have shown that convolutional networks can effectively classify sheep facial expressions with significant accuracy \cite{NOOR2020105528,7961768,10388128}. However, these methods have had limited application on farmland, and we propose that the primary challenge is not precision. A welfare determination about separation of an animal, the choice to contact a veterinarian, or management of a medication has costs and impact, and an estimated pain assessment provided without explanation offers the stockperson little justification for evaluation. What a practitioner needs is what the SPFES offers; a statement of which facial attributes are abnormal and how they vary.. Explainability is not only an optional enhancement but a fundamental requirement for the effective use of the innovation of system models to assist the health of sheep.

This motivates a distinctive system model point of view. If the clinical scale essentially provides a semantic identification of defined descriptors associated with anatomy-specific areas of study, then the most logical method to further explain a model's reasoning is to integrate those descriptors inside the inference process rather than justify after the fact. We implement a fixed vision-language model to encode the SPFES descriptors, allowing each identified face area to focus on these areas. The resultant attention is not a saliency heuristic evaluated afterwards; it is the process by which the representation of the region is derived, so the explanation and computation are the same objects. A shallow graph addressing the detected areas, weighted by anatomical proximity, enables the integration of information from the eyes, ears, and nose in a way reflecting the practical application of the scale as shown in Figure~\ref{fig:overlay}.

\section*{Nomenclature}
\label{sec:nomenclature}

\noindent\footnotesize
\begin{tabular}{@{}ll@{}}
\toprule
Symbol & Meaning \\
\midrule
$I$ & input image of a single sheep face \\
$N$ & number of facial regions detected in $I$ \\
$N_{\max}$ & maximum number of region nodes retained ($N_{\max}=9$) \\
$i, j$ & indices over region nodes, $i,j \in \{1,\dots,N\}$ \\
$b_i$ & bounding box of region $i$ \\
$r_i$ & anatomical type of region $i$ (eyes, ears or nose) \\
$\mathbf{x}_i$ & image crop of region $i$ \\
$f_\theta$ & convolutional region encoder with parameters $\theta$ \\
$\mathbf{h}_i \in \mathbb{R}^{d}$ & appearance embedding of region $i$ \\
$d$ & width of the region embedding ($d = 256$) \\
$\mathbf{A} \in \mathbb{R}^{N\times N}$ & weighted anatomical adjacency \\
$A_{ij}$ & edge weight between regions $i$ and $j$ \\
$w(\cdot,\cdot)$ & function assigning edge weight from region types \\
$\mathbf{I}$ & identity matrix (self-loops) \\
$\tilde{\mathbf{D}}$ & diagonal degree matrix of $\mathbf{A}+\mathbf{I}$ \\
$g(\mathbf{L})$ & symmetrically normalized graph filter \\
$\delta_{ij}$ & Kronecker delta, $1$ if $i=j$ else $0$ \\
$\mathcal{D}=\{d_1,\dots,d_C\}$ & set of SPFES clinical descriptors \\
$C$ & number of descriptors ($C = 12$) \\
$c$ & index over descriptors, $c \in \{1,\dots,C\}$ \\
$g_\phi$ & frozen vision--language text encoder \\
$\mathbf{t}_c \in \mathbb{R}^{d_t}$ & embedding of descriptor $d_c$ \\
$d_t$ & width of the text embedding ($d_t = 512$) \\
$\mathbf{W}_q,\mathbf{W}_k,\mathbf{W}_v$ & query, key and value projections \\
$m_{ic} \in \{0,1\}$ & anatomical mask (region--descriptor compatibility) \\
$\alpha_{ic}$ & attention of region $i$ on descriptor $c$ \\
$\mathbf{u}_i$ & clinical context vector attended by region $i$ \\
$\gamma_i \in (0,1)$ & language gate for region $i$ \\
$\mathbf{W}_g$ & gate projection \\
$\sigma(\cdot)$ & logistic sigmoid \\
$\odot$ & elementwise product \\
$[\,\cdot\,;\cdot\,]$ & vector concatenation \\
$\mathbf{z}_i$ & language-conditioned representation of region $i$ \\
$\mathbf{z}'_i$ & representation after graph message passing \\
$M_{ij}$ & message-passing coefficient from node $j$ to $i$ \\
$K$ & number of message-passing hops ($K = 1$) \\
$\mathbf{W}$ & message-passing weight matrix \\
$\mathbf{z}_G$ & graph-level (image-level) descriptor \\
$\mathbf{W}_0$ & frozen pretrained weight matrix \\
$\mathbf{A}_L,\mathbf{B}_L$ & low-rank LoRA factors \\
$r$ & LoRA rank ($r = 16$) \\
$\alpha_L$ & LoRA scaling ($\alpha_L = 32$) \\
$y \in \{0,1,2\}$ & image-level pain label \\
$\kappa$ & Cohen's kappa \\
\bottomrule
\end{tabular}
\normalsize

A second condition is determined by the area in which such a system would operate. Welfare surveillance is gradually expected on mobile and robotic platforms, portable devices at the race, stationary cameras at the water trough \cite{HITELMAN2022106713}, and unmanned aerial vehicles for flock-level monitoring \cite{10155900,SARWAR2021106219}. These systems are limited in memory, energy, and the capacity for on-site retraining. Therefore, we examine the model using low-rank adaptation (LoRA), maintaining the pretrained backbone in a predetermined state while just learning a rank-$r$ modification. This decreases the number of trainable parameters by a factor of $\times$, which is significant for on-device changes and for the practical issue of how a model is updated when encountering a new form, enclosure system, or camera.

Our third commitment concerns evaluation. Our previous work on this dataset \cite{11099391} addressed a related but different task: a YOLOv8-Nano detector predicts a pain level for each facial region, and a weighted graph aggregates
those per-region predictions into an image-level score, so the accuracy reported there reflects per-region detection quality combined with a fixed
aggregation rule. In the present work no per-region pain level is supplied at any stage; the model receives only region localization and type and must infer pain from appearance. The two settings therefore measure different quantities, and their numbers are not directly comparable. We evaluate all approaches, earlier formulations and the proposed models alike, under one methodology using
stratified five-fold cross-validation, a consistent training budget, and identical augmentation. This yields what is, to our knowledge, the first internally consistent benchmark on this dataset, and it produces a result we report without qualification: on region graphs of at most nine nodes, graph-based relational reasoning does not
improve pain discrimination over a well-trained region encoder with mean pooling. We regard this as useful for a literature that has applied graph models to small part graphs with limited ablation.

The contributions of this work are as follows.

\begin{itemize}
\item We propose a causal analysis of language-based explanation. We show that a system model using SPFES descriptors via residual cross-attention achieves explanations that are descriptive: removal of a detailed clinical descriptor affects the predicted logit by approximately $10^{-4}$, and the cue receiving the most attention correlates with the predicted pain level in $32.6\%$ regions. The attention maps appear logical, and the learning gate remains open, so normal interpretability evaluations do not indicate a problem.

\item We propose a concept-bottleneck formulation that removes the bypass and the demonstration that architecture alone is not enough. The classifier reads only a vector of SPFES concept scores, so the vocabulary is causally necessary, but without the per-region supervision that image-level pipelines discard, concept accuracy collapses from $0.801$ to $0.109$ while pain agreement is unchanged.
A bottleneck can be causally necessary and semantically meaningless at once, and no accuracy figure reveals the difference.

\item We provide evidence that the concepts are genuinely learned, together with the price of learning them. Minority SPFES states are recovered at $3.5$--$8.3\times$ their base rates, and the ear and eye severity orderings emerge without severity supervision, at a cost of $0.05$--$0.10$ in $\kappa$. We further quantify a trade-off internal to the design: relational message passing raises pain agreement by $0.078$ while lowering concept fidelity by $0.110$, because mixing neighboring regions makes a per-region explanation
partly about other regions.

\item We justify that a pooled concept accuracy is misleading under clinical class imbalance. Ours is $0.801$, below the $0.869$ of a constant majority-state predictor, while the model is nonetheless doing the clinically useful thing. Balanced recall against per-concept base rates should be reported in its place. Moreover, a corrected, cross-validated benchmark of seven methods under one matched protocol, to our knowledge, is the first internally consistent evaluation on this dataset including the negative result that relational reasoning over graphs of at most nine anatomical nodes does not improve pain discrimination over a well-trained region encoder.
\end{itemize}

The remaining sections of this paper are structured as follows. Section~\ref{sec:related} reviews related work on sheep facial analysis, graph neural networks for behavioral biometrics, vision language models and interpretability, parameter-efficient adaptation, and deployment. Section~\ref{sec:method} presents the complete system model of the language-grounded explanations with interpretability to monitor sheep facial pain. Section~\ref{sec:setup} describes the experimental evaluation, including the dataset and protocols. Section~\ref{sec:discussion} presents when relational reasoning does not help in small anatomical graphs and how LoRA brings efficiency to the field deployment. While section~\ref{sec:limitations} explains the limitations, and in the last section~\ref{sec:conclusion} defines the conclusion.  

\section{Related Work}
\label{sec:related}

\paragraph{Facial pain assessment in sheep}
The SPFES \cite{MCLENNAN201619} shown that stress in sheep causes predictable, observable changes in face morphology, confirmed by footrot and mastitis. The computational studies followed with two general trajectories. The first approach considers pain as a form of visual category, like Noor et al. \cite{NOOR2020105528} developed convolutional models that differentiated between normal and abnormal sheep faces, and following systems expanded this to include landmark localization and action-unit assessment \cite{11099391,7961768,10388128,7477733}. The second direction corresponds to detection and identification rather than distress for animal based facial recognition and biometric identification, which possess improved precision with optimized architectures suitable for agricultural utilization \cite{HITELMAN2022106713,ZHANG2022107452,LI2023107651,ani13111824,YUAN2025101362}. Our concern is the first line, and specifically with what such systems communicate to the user rather than how accurately they score.

\paragraph{Graph reasoning over facial parts}
Representing a face as a graph of parts is attractive when the clinical instrument is itself organized by region. Graph convolutional networks \cite{2016arXiv160902907K} and graph attention networks \cite{veličković2018graph} provide the standard machinery, and graph parsing has been applied to structured visual reasoning \cite{Qi2018LearningHI}. Noor et al. \cite{11099391} work applied a weighted graph formulation to sheep facial pain. A relevant caution comes from theory to bound generalization for deep graph convolutional networks, which might change into low precision with depth \cite{11186137,10.1145/3292500.3330956}, and node representations may lose expressive power as layers accumulate \cite{oono2020graph}. On graphs of the size considered here, at most nine nodes spanning three region types, these results show depth is not merely unhelpful but actively harmful to accuracy, motivating the single-hop design we adopt and consistent with what we observe empirically.

\paragraph{Vision language models and interpretability}
Contrastive image2text pretraining \cite{pmlr-v139-radford21a} generates a unified embedding that allows for direct comparison between natural-language expressions and visual information. This aligns with clinical scales expressed in language: instead of interpreting an ambiguous correlation from pixels to a pain score, a model may be inquired on the degree to which an area fits each given descriptor. Previous research on graph convolution indicates how important node attributes enhance linear separability \cite{2021arXiv210206966B}, which is the function that the fixed language embeddings serve in our framework. We maintain the text encoder in a static state, ensuring that the language prior introduces no trainable parameters, with its semantics derived from pretraining rather than adapted to a limited dataset.

\paragraph{Parameter-efficient adaptation and deployment}
Low-rank adaptation (LoRA) fixes a pretrained network and learns a low-rank adaptation for particular weight matrices, minimizing trainable parameters while maintaining the pretrained representation. For agricultural applications, this is more than just convenience. UAV-assisted livestock surveillance is a prominent field, and the critical need for lightweight on-device models is frequently highlighted in the sheep-monitoring research \cite{ani13111824,YUAN2025101362,agriculture14030468}. A model that allows for adaptation to a new flock or camera by modifying only a tiny portion of its weights is far simpler to implement and sustain than one that requires comprehensive fine-tuning.

This article extends our earlier conference paper \cite{11099391} in a different domain, from which the dataset and the anatomical-graph formulation are taken. That work established a sheep facial landmark dataset under SPFES parameters and proposed a weighted graph neural network that clusters detected facial parts and aggregates their pain levels into an image-level score, reporting $92.71\%$ accuracy with a YOLOv8-Nano detector. The present article differs in three respects. First, the task is different; there, per-region pain levels predicted by the detector are inputs to the graph, whereas here the model is given only region localization and type and must infer pain from appearance, so the two sets of numbers quantify different things, and we do not compare them directly. Second, the contribution is interpretability rather than aggregation: we test whether language-grounded explanations are causally connected to the prediction, find that they are not, and replace the architecture with a concept bottleneck whose clinical vocabulary is load-bearing by construction. Third, the evaluation is expanded from a single split to stratified five-fold
cross-validation across seven methods under one matched protocol, with concept-level as well as pain-level metrics. None of the experiments reported
here appear in \cite{11099391}.
\section{Methodology}
\label{sec:method}

The system model is developed through four different stages. An image shrinks down to face areas; each area is encoded and associated with the SPFES clinical vocabulary; the regions exchange information via an anatomical graph; and a pooled descriptor is divided into categories while the corresponding alignment weights serve as the explanation. 

\subsection{Region graph construction}
\label{sub:graph}

Each image $I$ is first reduced to a set of facial regions. A detector trained on the dataset's annotation scheme localizes instances of $3$ anatomical region types, yielding $N \le N_{\max}$ regions with bounding boxes $b_i$ and region types
$r_i \in \{\textrm{eyes}, \textrm{ears}, \textrm{nose}\}$, for $i = 1,\dots,N$. Here $N$ is the number of regions actually detected in the image, and $N_{\max} = 9$ is a fixed capacity chosen to accommodate every image in the corpus; images with fewer regions are padded, and the padding is masked out at every subsequent stage. Regions annotated as not classifiable are retained as graph nodes, since they carry valid appearance evidence but do not contribute to the image-level label. The regions form the nodes of an undirected graph whose edges encode anatomical coupling. Let $\mathbf{A} \in \mathbb{R}^{N \times N}$ be the weighted adjacency matrix, whose entry $A_{ij}$ is the coupling strength between regions $i$ and $j$. We set $A_{ij} = w(r_i, r_j)$, where $w(\cdot,\cdot)$ returns a full weight for distinct region types and a reduced weight when $r_i = r_j$.

Concretely, $w(r_i,r_j)=1$ for distinct region types and $w(r_i,r_j)=0.5$ when $r_i=r_j$. The reduction encodes a conditional-independence assumption; two detections of the same anatomical type constitute repeated observations of
a single underlying state, whereas an ear and a nose report on distinct components of the expression, so a same-type edge is assigned proportionally less influence in the message-passing step. The value was fixed a priori and held constant across every method that employs a weighted adjacency, including re-implementation of the formulation of \cite{11099391}, so that no comparisons in evaluations are confounded by it; it was not, however, subjected to ablation, and its contribution to the reported agreement therefore remains unquantified. Message passing uses the symmetrically normalized filter with self-loops,
\begin{equation}
\label{eq:filter}
g(\mathbf{L}) = \tilde{\mathbf{D}}^{-1/2}
\big(\mathbf{A} + \mathbf{I}\big)
\tilde{\mathbf{D}}^{-1/2},
\qquad
\tilde{D}_{ii} = \sum_{j=1}^{N} \big(A_{ij} + \delta_{ij}\big),
\end{equation}
in which $\mathbf{I}$ is the $N \times N$ identity matrix adding a self-loop to every node so that a region retains its own evidence; $\tilde{\mathbf{D}}$ is the diagonal degree matrix whose $i$-th diagonal entry $\tilde{D}_{ii}$ sums the weights incident on node $i$ including its self-loop; $\delta_{ij}$ is the Kronecker delta, equal to one when $i = j$ and zero otherwise; and $g(\mathbf{L})$ denotes the resulting normalized filter, written as a function of the graph Laplacian by convention. The two-sided normalization by $\tilde{\mathbf{D}}^{-1/2}$ bounds the spectral norm of $g(\mathbf{L})$ independently of graph size, so the per-layer contraction factor that enters generalization bounds for graph convolutional networks \cite{11186137} does not grow with $N$. This matters here because $N$ varies between images.

\subsection{Region encoding}
\label{sub:encoder}

Each region is cropped with a small context margin and encoded by a convolutional backbone $f_\theta$ shared across regions and across region types,
\begin{equation}
\label{eq:encoder}
\mathbf{h}_i = f_\theta(\mathbf{x}_i) \in \mathbb{R}^{d},
\qquad i = 1,\dots,N,
\end{equation}
where $\mathbf{x}_i$ is the crop of region $i$, $\theta$ are the backbone parameters, $\mathbf{h}_i$ is the resulting appearance
embedding and $d = 256$ is the width. Sharing $f_\theta$ across region types is deliberate with small training samples; per-region encoders
would triple the parameter count while dividing the effective sample size for each. We use a ResNet-18 initialized from ImageNet weights, with the stem and first residual stage frozen, since on a corpus of this size adapting the earliest filters yields no benefit and increases variance.

\subsection{SPFES descriptor cross-attention}
\label{sub:xattn}

The clinical scale is a list of phrases, and we use them directly. Let $\mathcal{D} = \{d_1, \dots, d_C\}$ denote the SPFES descriptors, where each $d_c$ is a short clinical phrase associated with one region type, for example, "orbital tightening around the eye, ears rotated backward, a V-shaped, and tense nose profile," together with a relaxed reference phrase for each region. With four descriptors per region type across three types, $C = 12$. A frozen text encoder $g_\phi$ from a contrastively pretrained vision language model \cite{pmlr-v139-radford21a} maps each phrase to an embedding,
\begin{equation}
\label{eq:textemb}
\mathbf{t}_c = g_\phi(d_c) \in \mathbb{R}^{d_t},
\qquad c = 1,\dots,C,
\end{equation}
where $\phi$ are the (fixed) text-encoder parameters and $d_t = 512$ is the embedding width. These are computed once and held constant. Because $g_\phi$ is never updated, the language prior contributes no trainable parameters, and its semantics are determined by large-scale pretraining
rather than being fitted to a small set of sheep images. Region $i$ attends over the descriptor set with queries derived from the region and keys and values from the frozen text embeddings,
\begin{equation}
\label{eq:attn}
\alpha_{ic} =
\frac{\exp\!\big(m_{ic}\,\langle \mathbf{W}_q \mathbf{h}_i,\,
\mathbf{W}_k \mathbf{t}_c \rangle / \sqrt{d}\big)}
{\sum_{c'=1}^{C} \exp\!\big(m_{ic'}\,\langle \mathbf{W}_q \mathbf{h}_i,\,
\mathbf{W}_k \mathbf{t}_{c'} \rangle / \sqrt{d}\big)}.
\end{equation}

In \eqref{eq:attn}, $\mathbf{W}_q$ projects the region embedding $\mathbf{h}_i$ into the query space and $\mathbf{W}_k$ projects the text embedding $\mathbf{t}_c$ into the matching key space, both into $\mathbb{R}^{d}$; $\langle \cdot,\cdot \rangle$ is the inner product; division by $\sqrt{d}$ is the standard scaling that keeps the logits in a range where the softmax is not saturated; and $\alpha_{ic} \in (0,1)$ with $\sum_c \alpha_{ic} = 1$ is the attention region $i$ places on descriptor $c$. The binary mask $m_{ic}$ equals one when descriptor $c$ belongs to the region type $r_i$ and zero otherwise, implemented by setting the masked logits to $-\infty$ before the softmax so that an eye region can attend only to eye descriptors. This mask is what makes the attention map clinically readable. Without it, attention mass could spread across anatomically irrelevant phrases and the weight $\alpha_{ic}$ would no longer admit the reading ``the degree to which region $i$ was compared against cue $d_c$''. With it, the row of $\alpha$ for a given region is a distribution over exactly the cues a clinician would consult for that region. The attended clinical context is the corresponding convex combination of projected descriptor embeddings,
\begin{equation}
\label{eq:context}
\mathbf{u}_i = \sum_{c=1}^{C} \alpha_{ic}\,\mathbf{W}_v \mathbf{t}_c
\in \mathbb{R}^{d},
\end{equation}
where $\mathbf{W}_v$ is the value projection. The context is injected into the region representation through a learned gate,
\begin{equation}
\label{eq:gate}
\mathbf{z}_i = \mathbf{h}_i + \gamma_i \odot \mathbf{u}_i,
\qquad
\gamma_i = \sigma\!\big(\mathbf{W}_g [\mathbf{h}_i ; \mathbf{u}_i]\big),
\end{equation}
in which $[\mathbf{h}_i ; \mathbf{u}_i] \in \mathbb{R}^{2d}$ is the concatenation of the appearance embedding and the clinical context, $\mathbf{W}_g$ is the gate projection, $\sigma$ is the logistic sigmoid constraining $\gamma_i$ to $(0,1)$, $\odot$ is the elementwise product, and $\mathbf{z}_i$ is the language-conditioned region representation. The gate makes the contribution of the language prior identifiable rather than assumed. If the clinical descriptors carried no usable signal, the loss would be minimized by driving $\gamma_i \to 0$, at which point \eqref{eq:gate} reduces to $\mathbf{z}_i = \mathbf{h}_i$ and the model becomes its appearance-only counterpart. The learned gate values are therefore a direct measurement of how much the model elects to rely on the clinical vocabulary, and we report them in Section~\ref{sec:results} as a diagnostic that an ablation alone cannot provide.

\subsection{Shallow relational reasoning and readout}
\label{sub:reasoning}

A single message-passing hop over the region graph produces
\begin{equation}
\label{eq:hop}
\mathbf{z}'_i = \tanh\!\Big(\textstyle\sum_{j=1}^{N} M_{ij}\,
\mathbf{W}\mathbf{z}_j\Big),
\end{equation}
where $M_{ij}$ is the coefficient with which node $j$ contributes to node $i$either the fixed entry $[g(\mathbf{L})]_{ij}$ of \eqref{eq:filter} or a learned attention over the same support $\mathbf{W}$ is a shared linear map applied to every node and $\mathbf{z}'_i$ is the updated representation. We use $\tanh$ rather
than a rectifier because it is zero-centered and Lipschitz, matching the activation assumptions under which the stability results of \cite{11186137} are stated. We set the number of hops $K = 1$. This is a design choice rather than a tuned hyperparameter. With at most $9$ nodes and a dense anatomical adjacency, a single hop already gives every node access to the whole face, so a second hop adds little reachability. Meanwhile, each additional layer multiplies the depth-dependent term in the generalization bound \cite{11186137} and accelerates the loss of node distinguishability \cite{oono2020graph}. The graph-level descriptor is obtained by masked mean pooling over valid nodes,
\begin{equation}
\label{eq:readout}
\mathbf{z}_G = \frac{1}{N}\sum_{i=1}^{N} \mathbf{z}'_i,
\end{equation}
and classified by a linear layer producing logits over the three pain levels $y \in \{0,1,2\}$. The mean in \eqref{eq:readout} is taken over valid/unpadded nodes only.

\subsection{Parameter-efficient adaptation (LoRA)}
\label{sub:lora}

For deployment we replace full fine-tuning of the backbone with LoRA. Each adapted convolution with pretrained weight $\mathbf{W}_0$ is frozen and augmented with a trainable low-rank update,
\begin{equation}
\label{eq:lora}
\mathbf{W} = \mathbf{W}_0 + \frac{\alpha_L}{r}\,
\mathbf{B}_L\mathbf{A}_L,
\qquad
\mathbf{A}_L \in \mathbb{R}^{r \times d_{\mathrm{in}}},\;
\mathbf{B}_L \in \mathbb{R}^{d_{\mathrm{out}} \times r},
\end{equation}
where $d_{\mathrm{in}}$ and $d_{\mathrm{out}}$ are the input and output widths of the layer, $r \ll \min(d_{\mathrm{in}}, d_{\mathrm{out}})$ is
the rank of the update, $\mathbf{A}_L$ and $\mathbf{B}_L$ are the two trainable factors, and $\alpha_L / r$ is a scaling that keeps the magnitude of the update comparable as $r$ varies. Only $\mathbf{A}_L$ and $\mathbf{B}_L$ receive gradients with $\mathbf{W}_0$ fixed. $\mathbf{B}_L$ is initialized to zero so that $\mathbf{W} = \mathbf{W}_0$ at the first step and adaptation begins exactly at the pretrained solution. We adapt the seventeen convolutions
of the backbone whose channel widths exceed a minimum threshold, using
$r = 16$ and $\alpha_L = 32$.

Two properties of this construction matter on a corpus of this size. First, the adapted weight remains within a bounded distance of the pretrained one, with the update $\mathbf{B}_L\mathbf{A}_L$ has rank at most $r$ and its norm is controlled by the norms of the two factors, which weight decay penalizes directly. Generalization bounds for the composition of a feature extractor and a graph stage depend on the norm of the trainable parameters \cite{11186137,10.1145/3292500.3330956}, and LoRA constrains that norm structurally rather than by regularization alone. Informally, full fine-tuning is free to move the representation anywhere in weight space and must be prevented from doing so by augmentation and early stopping, whereas LoRA cannot move it far in the first place. Second, the choice of which layers to adapt is not neutral. We adapt convolutions whose channel widths exceed a threshold, which in practice means the later residual stages, leaving the stem and earliest filters frozen. Early convolutional filters encode generic edge and texture statistics that transfer across domains essentially unchanged; the later stages encode the composite structures that differ between ImageNet objects and sheep facial regions. Spending a limited adaptation budget on the layers that actually need to change is what allows a $9.6\%$ trainable fraction to remain competitive.

\subsection{Testing whether the explanation explains}
\label{sub:attribution}

The construction of SPFES descriptor cross-attention and shallow relational reasoning are the ones an interpretability-minded reader would expect, and its explanations look convincing. While the attention map is cleanly organized by region, the learned gate stays open, and a fluent sentence naming clinical cues falls out directly. We nevertheless think such a construction should be tested rather than assumed, and the test is simple. For region $i$ and descriptor $c$ we recompute the forward pass with $(i, c)$ removed from the attention support and record the change in the logit of the
predicted class, 
\begin{equation}
\label{eq:attrib}
a_{ic} \;=\; \ell_{\hat{y}}(\text{full}) \;-\;
\ell_{\hat{y}}(\text{without descriptor } c \text{ at region } i),
\end{equation}
where $\ell_{\hat{y}}$ is the logit of the predicted class $\hat{y}$. A positive $a_{ic}$ means the descriptor supported the prediction and a
negative one that it argued against it. Unlike the attention weight, $a_{ic}$ is obtained by intervening on the computation rather than by reading a quantity off it, so it measures what the descriptor \emph{does} rather than how strongly it was consulted. The cost is $C$ additional forward passes per image, which is negligible. We report in section~\ref{sub:attrib-results} the outcome, and it is not the one the attention maps suggest.

\subsection{Concept-bottleneck formulation}
\label{sub:cbm}

The failure diagnosed in section~\ref{sub:attrib-results} is architectural rather than incidental. In \eqref{eq:gate} the clinical context enters as a residual added to an appearance representation that already suffices for the task, so the optimizer is under no pressure to use it; a gate that stays open is cheap when the vector it admits is nearly constant. Any fix that leaves the appearance pathway intact leaves the incentive intact. We therefore remove the bypass, following the concept-bottleneck idea \cite{koh2020cbm} but adapting it to a clinical scale whose concepts are both
named and, in this dataset, individually annotated. Let $\mathcal{K} = \{k_1,\dots,k_M\}$ be the set of SPFES states be the annotation vocabulary. It distinguishes ear flat, ear rotated, ear flipped, eye open, eye partly closed, nose shallow-U, nose shallow-V, and nose extended-V, so $M = 8$ each belongs to one region type and each carries a clinical phrase. Writing $\mathbf{q}_i = W_t\mathbf{h}_i$ for the region embedding projected into the frozen text space and $\mathbf{t}_m$ for the frozen embedding of state $m$, the per-region concept distribution is
\begin{equation}
\label{eq:concept}
p_{im} = \frac{\exp\!\big(\tau\,
\hat{\mathbf{q}}_i^{\top}\hat{\mathbf{t}}_m\big)\,\mu_{im}}
{\sum_{m'} \exp\!\big(\tau\,
\hat{\mathbf{q}}_i^{\top}\hat{\mathbf{t}}_{m'}\big)\,\mu_{im'}},
\end{equation}
in which $\hat{\cdot}$ denotes $\ell_2$ normalization, $\tau$ is a learned temperature, and $\mu_{im}\in\{0,1\}$ masks states belonging to a different region type so that an ear region is scored only against ear states. The image-level concept vector takes, for each state, the strongest evidence over the regions that could express it,
\begin{equation}
\label{eq:cvec}
z_m = \max_{i \,:\, \mu_{im}=1} p_{im},
\end{equation}
the maximum rather than a mean so that one affected ear is not diluted by an unaffected one. Pain is then predicted by a linear map from $\mathbf{z} \in \mathbb{R}^{M}$ alone,
\begin{equation}
\label{eq:cbmhead}
\boldsymbol{\ell} = W_c \mathbf{z} + \mathbf{b},
\qquad W_c \in \mathbb{R}^{3 \times M}.
\end{equation}
No path from appearance to the decision avoids $\mathbf{z}$, so the near-zero attribution of Section~\ref{sub:attrib-results} cannot recur: the concepts are causally necessary by construction. Two properties keep the bottleneck explainable. The coordinates of $\mathbf{z}$ are alignments to \emph{frozen} text embeddings, so they retain linguistic meaning rather than drifting into an arbitrary eight-dimensional code. And they \emph{directly supervised} the detection classes and named the SPFES state of each region (ear flat, eyes partly closed, nose extended in a V shape, and so on), giving a per-region label that image-level pipelines discard. Training
minimizes
\begin{equation}
\label{eq:cbmloss}
\mathcal{L} = \mathrm{CE}\big(\boldsymbol{\ell}, y\big)
\;+\; \lambda \!\!\sum_{i \,:\, k_i \neq \varnothing}\!\!
\mathrm{CE}\big(p_{i\cdot}, k_i\big),
\end{equation}
with $k_i$ the annotated state of region $i$, the sum running over assessable regions only, and $\lambda = 1$. Note the supervision constrains
concept \emph{identity} and never severity; the ordering of states within a region is therefore free to be recovered, or not, from data, which makes it usable. Because \eqref{eq:cbmhead} is linear in $\mathbf{z}$, the weight matrix $W_c$ is a global explanation: the contribution of every SPFES state to every pain level, in logit units, is readable directly. One caveat governs its interpretation. The scores of a region's states sum to one by \eqref{eq:concept}, so the columns of $W_c$ within a region are linearly dependent, and the weights are identified only up to an additive constant per region. Raw signs are therefore meaningless; we center within each region
before reporting, which removes the unidentified constant and leaves the differences, which are what carry information.

\section{Experimental Setup}
\label{sec:setup}
\subsection{Dataset}

We use the sheep facial pain dataset introduced in \cite{11099391}, annotated according to the SPFES \cite{MCLENNAN201619}. Images are ground-level, close-range photographs of individual animals. Region annotations span ten classes,
encoding region type together with an ordinal pain level. After parsing, dataset images carry at least one assessable region. The image-level label is the maximum pain level over assessable regions, giving three classes distributed as $334$ no pain, $126$ moderate pain, and $24$ severe pain. This imbalance, and in particular the scarcity of severe-pain examples, is a substantive limitation discussed in Section~\ref{sec:limitations}. Only region type and localization are supplied to the model; the pain sub-label attached to each detected region is never used as an input feature. This differs from the study of \cite{11099391}, in which per-region pain levels predicted by the detector are aggregated by the graph, and it is why the absolute agreement values reported here are lower than, and not comparable to,
those previously published.

\subsection{Protocol}

Because the dataset's held-out test split contains only limited images, at which size a single Cohen's $\kappa$ is dominated by sampling noise, we report stratified five-fold cross-validation over the pooled corpus. In each fold, one partition serves as a test, one as validation for model selection, and the remainder for training. We report the mean $\pm$ standard deviation across folds. Cohen's $\kappa$ is the primary metric, and accuracy is reported alongside. The choice is consequential given the label distribution. A model that predicts no pain for every image achieves an accuracy of $0.69$ on this corpus while providing no clinical value whatsoever; the same model scores
$\kappa = 0$. Because $\kappa$ subtracts the agreement expected from the marginal distributions alone, it is the metric that distinguishes a system that has learned something about pain from one that has learned the prior. Every accuracy figure in this paper should be read against that $0.69$ floor rather than against $0.33$. Every method in experiments uses an identical protocol with the same folds, the same region encoder and initialization, the same optimizer (AdamW, learning rate $3\times10^{-4}$, backbone learning-rate multiplier $0.1$, weight decay $10^{-4}$), the same cosine schedule and mixed-precision training, the same augmentation (horizontal flip, color jitter, small affine perturbation, random erasing), the same class-balanced sampling and label smoothing of $0.05$, and the same budget of $150$ epochs on validation $\kappa$ at patience $35$. The frozen text encoder is CLIP ViT-B/32. The LoRA configuration uses a higher learning rate ($10^{-3}$) and the same extended schedule.

\section{Results}
\label{sec:results}
\subsection{Comparison under a matched protocol}

Table~\ref{tab:sota} reports all methods under the matched protocol. Among the models that make no attempt at explanation, a
graph-free region encoder with mean pooling is strongest at $\kappa = 0.609 \pm 0.046$, followed by a max-pooling variant at $0.572 \pm 0.094$. The graph-based methods span $0.536$ to $0.578$: our prior weighted-graph formulation \cite{11099391}, re-implemented under the proposed protocol, reaches $0.567 \pm 0.092$, a graph attention network
\cite{veličković2018graph} $0.548 \pm 0.057$, and a graph convolutional network
\cite{2016arXiv160902907K} $0.536 \pm 0.047$. The residual SPFES attention model attains $0.578 \pm 0.084$, the highest of the graph-based group and the concept bottleneck $0.524 \pm 0.077$. The standard deviations overlap substantially throughout, and we do not claim that these differences are individually significant. The model with the best agreement among the language-grounded variants is precisely the one whose explanations do not survive testing. 

\begin{table*}[!t]
\centering
\caption{Five-fold cross-validated comparison under a matched protocol (identical folds, encoder, optimizer, augmentation, and $150$-epoch budget).
Mean $\pm$ standard deviation across folds. Only the concept bottleneck provides an explanation that survives a causal test; the models above it
either offer none or offer one that does not hold up
(Section~\ref{sub:attrib-results}).}
\label{tab:sota}
\small
\begin{tabular}{lccc}
\toprule
Method & Cohen's $\kappa$ & Accuracy & Trainable params \\
\midrule
Region encoder + MLP (no graph)      & $0.609 \pm 0.046$ & $0.831 \pm 0.023$ & $11.2$M \\
Max-pool readout (no graph)          & $0.572 \pm 0.094$ & $0.805 \pm 0.045$ & $11.2$M \\
2D Weighted GNN \cite{11099391} ($K{=}2$)$^{\dagger}$ & $0.567 \pm 0.092$ & $0.808 \pm 0.045$ & $11.3$M \\
GAT \cite{veličković2018graph} ($K{=}2$) & $0.548 \pm 0.057$ & $0.802 \pm 0.023$ & $11.3$M \\
GCN \cite{2016arXiv160902907K} ($K{=}2$)     & $0.536 \pm 0.047$ & $0.793 \pm 0.026$ & $11.3$M \\
Residual SPFES attention ($K{=}1$)$^{\ddagger}$ & $0.578 \pm 0.084$ & $0.814 \pm 0.030$ & $11.5$M \\
\midrule
\textbf{SPFES concept bottleneck} & $\mathbf{0.524 \pm 0.077}$ & $0.804 \pm 0.034$ & $11.3$M \\
Residual attention + LoRA ($r{=}16$) & $0.529 \pm 0.091$ & $0.799 \pm 0.038$ & $\mathbf{1.18}$\textbf{M} \\
\bottomrule
\end{tabular}
\\[2pt]
\footnotesize
$^{\dagger}$The weighted-graph formulation of \cite{11099391} adapted to the present task and trained under this protocol. In \cite{11099391} the graph aggregates per-region pain levels supplied by the detector; here it receives no pain input, so this row is an adaptation rather than a reproduction and is not comparable to the accuracy published there. $^{\ddagger}$Highest agreement among the language-grounded variants, but its explanations are inert (Section~\ref{sub:attrib-results}); it is listed for completeness rather than as a recommended model. Only the concept
bottleneck supports a verified explanation.
\end{table*}


\subsection{Parameter-efficient adaptation}

Adapting the model through LoRA reduces trainable parameters from $11.5$M to $1.18$M, or $9.6\%$ of the network, at a cost of $0.049$ in $\kappa$ within one standard deviation of the fully fine-tuned model. Accuracy is essentially preserved ($0.799$ versus $0.814$). Two observations from the training dynamics are worth recording. LoRA converges considerably more slowly under a $60$-epoch budget with rank $8$ it reached only $\kappa = 0.391 \pm 0.078$, with four of five folds
terminating early while validation agreement was still rising. Extending the schedule to $150$ epochs and the rank to $16$ recovered $0.529 \pm 0.091$. Practitioners adopting parameter-efficient adaptation on small agricultural datasets should budget substantially more epochs than full fine-tuning requires; a short schedule will understate the method badly. This is a practical finding rather than a theoretical one, but it is the kind of detail that determines whether a method is adopted or discarded after a first unsuccessful trial.

\subsection{The language pathway does not contribute}
\label{sub:attrib-results}

\begin{figure*}[!t]
\centering
\includegraphics[width=\linewidth]{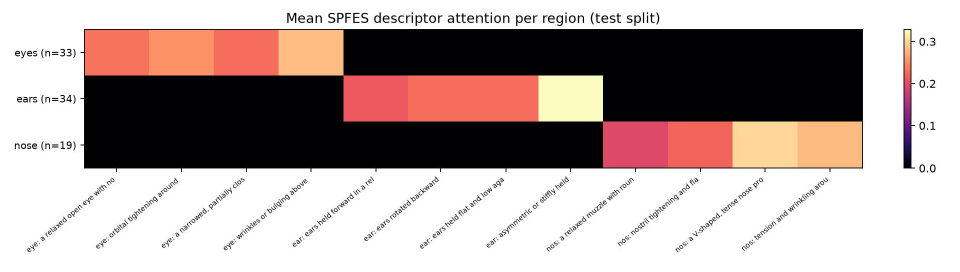}
\caption{Mean SPFES descriptor attention per region in the residual-attention model. The block structure is imposed by the anatomical mask and is therefore fixed rather than learned; within each block the weights sit near $0.25$, the
uniform value over four descriptors. The map looks informative and is not; ablating any descriptor changes the prediction by about $10^{-4}$.}
\label{fig:attention}
\end{figure*}

\begin{figure}[!t]
\centering
\includegraphics[width=0.85\linewidth]{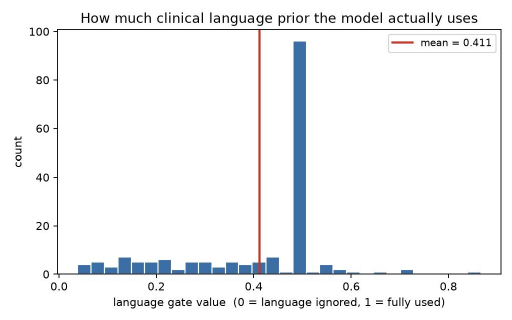}
\caption{Learned language gate $\gamma_i$ of \eqref{eq:gate} over test regions. A gate near zero would signal that the clinical prior is ignored;
the observed mean of $0.411$ suggests the opposite. The intervention test shows the gate stays open because the vector it admits is nearly constant
across images, not because it is informative.}
\label{fig:gate}
\end{figure}

The residual cross-attention model presents every outward sign of being language-grounded. Attention is cleanly organized by region (Figure~\ref{fig:attention}); within each region it is graded across the descriptors rather than collapsed onto one; and the learned gate of \eqref{eq:gate} does not close, sitting at a mean of $0.411$ with a pronounced mode near $0.5$ (Figure~\ref{fig:gate}). On the usual
diagnostics, the clinical vocabulary appears to be in use. The ablation test of \eqref{eq:attrib} contradicts this flatly. Removing an entire SPFES descriptor from a region changes the predicted-class logit by about $10^{-4}$, against logits of order unity to 10. The effect is not small but absent. Consistently, the descriptor with the largest attribution agrees in polarity with the predicted pain level in $32.6\%$ of regions ($28$ of $86$), which is worse than chance for a binary agreement.

The mechanism is visible in the attention itself once one looks for it. The
anatomical mask restricts each region to four descriptors, and the resulting softmax sits near $0.25$ on each of them. Almost all the structure apparent in Figure~\ref{fig:attention} is contributed by the mask, which is fixed; the \emph{within-region} weights, which are the only part that could carry per-image clinical information, are close to uniform. The attended context $\mathbf{u}_i$ of \eqref{eq:context} is therefore approximately the mean of a region's descriptor embeddings, nearly the same vector for every image, and the gate stays open because admitting a near-constant vector is harmless rather than because the vector is informative. The corresponding ablation row in Table~\ref{tab:ablation} agrees: removing the cross-attention entirely costs $0.010$ $\kappa$, well inside the fold standard deviation. We report this at length because none of it is visible in the artefacts an interpretability paper normally shows. The attention heat map, the open gate,
and the fluent generated sentence are all consistent with a model that reasons over clinical language, and all three were produced by a model that does not. The lesson we draw is that attention-based grounding claims should be accompanied by an intervention test and that the test is cheap.

\subsection{Concept bottleneck: accuracy}
\label{sub:cbm-acc}

Constraining the decision to pass through the SPFES concept vector costs accuracy, as expected when a three-way decision must be carried by eight
numbers. Under the identical five-fold protocol, the bottleneck attains $\kappa = 0.524 \pm 0.077$ with accuracy $0.804 \pm 0.034$ (Table~\ref{tab:sota}), against $0.578 \pm 0.084$ for the residual-attention model, and $0.609 \pm 0.046$ for the graph-free encoder that offers no explanation at all. The gap is $0.054$ and $0.085$ respectively, roughly one standard deviation, and the bottleneck is statistically level with GCN, GAT, and the LoRA variant. We regard this as the paper's central quantity. Interpretability is often
argued for qualitatively; here its price is measured, on one protocol, at
about $0.05$--$0.09$ $\kappa$.

\subsection{Concept bottleneck: Are the concepts learned?}
\label{sub:cbm-concepts}
Pain agreement cannot answer this question, so we evaluate the bottleneck coordinates directly against the per-region state annotations. The pooled concept accuracy is $0.801 \pm 0.030$. Taken alone this looks encouraging and is the number a concept-bottleneck study would ordinarily
report. The annotated states are severely imbalanced over the $1{,}941$ annotated regions; ear flat and eye open alone account for $72\%$, and within each region type, the majority of the state covers $0.858$ to $0.892$ of instances. A constant predictor that always names the majority state of a region therefore attains $0.869$, and the model's $0.801$ is \emph{below} that floor. The pooled concept nevertheless misdescribes what the model does, as Table~\ref{tab:concepts} shows. The two overwhelming majority states are recovered at $0.975$, slightly above, their base rates. The minority states the ones that indicate pain are recovered far above theirs; ear rotated at $0.525$ against a base rate of $0.091$ ($5.8\times$), ear flipped at $0.324$ against $0.039$ ($8.3\times$), eye partly closed at $0.587$ against $0.142$ ($4.1\times$), and nose shallow-V at $0.371$ against $0.105$ ($3.5\times$). Balanced recall across the supported states is $0.563 \pm 0.071$, against $0.429$ for the majority predictor, which recovers only the three majority states and none of the rest.

The model is therefore trading a little accuracy on two states that dominate the corpus in order to detect the rare states that carry clinical meaning, which is the behavior one wants from a pain detector and precisely what the pooled concept conceals. We draw a methodological conclusion under the class imbalance typical of clinical annotation. Pooled concept accuracy is not evidence of concept learning, and per-concept recall against base rates,
summarized by balanced recall, should be reported instead. Two states cannot support any claim. Nose shallow-U, though a majority state, is recovered at only $0.243$, indicating the nose head is close to
uninformative. Nose extended-V, clinically the most important nose state, occurs once in the entire corpus and is never recovered. The nose block is thus fitted on almost no evidence, which explains its behavior
in Section~\ref{sub:ordering}.

\begin{table*}[!t]
\centering
\caption{Per-concept recovery, pooled over the five folds. ``Share'' is the concept's frequency within its own region type, which is the rate a constant
predictor achieves for it; ``lift'' is recall divided by share. Minority states those indicating pain are recovered well above their base rates,
which the pooled accuracy of $0.801$ conceals.}
\label{tab:concepts}
\small
\begin{tabular}{llrrrrl}
\toprule
Concept & Region & $n$ & Share & Recall & Lift & \\
\midrule
ear flat        & ears & 762 & 0.870 & 0.975 & $1.1\times$ & majority state \\
ear rotated     & ears &  80 & 0.091 & 0.525 & $\mathbf{5.8\times}$ & above base rate \\
ear flipped     & ears &  34 & 0.039 & 0.324 & $\mathbf{8.3\times}$ & above base rate \\
eye open        & eyes & 629 & 0.858 & 0.975 & $1.1\times$ & majority state \\
eye partly closed & eyes & 104 & 0.142 & 0.587 & $\mathbf{4.1\times}$ & above base rate \\
nose shallow-U  & nose & 296 & 0.892 & 0.243 & $0.3\times$ & poorly recovered \\
nose shallow-V  & nose &  35 & 0.105 & 0.371 & $\mathbf{3.5\times}$ & above base rate \\
nose extended-V & nose &   1 & 0.003 & 0.000 &  & support too small \\
\midrule
\multicolumn{6}{l}{pooled concept accuracy} & $0.801 \pm 0.030$ \\
\multicolumn{6}{l}{majority-state baseline} & $0.869 \pm 0.003$ \\
\multicolumn{6}{l}{balanced (macro) recall} & $\mathbf{0.563 \pm 0.071}$ \\
\multicolumn{6}{l}{majority predictor, balanced recall} & $0.429$ \\
\bottomrule
\end{tabular}
\end{table*}

\subsection{Qualitative explanations, and a leak made visible}
\label{sub:cbm-qual}
\begin{figure*}[!t]
\centering
\includegraphics[width=0.62\linewidth]{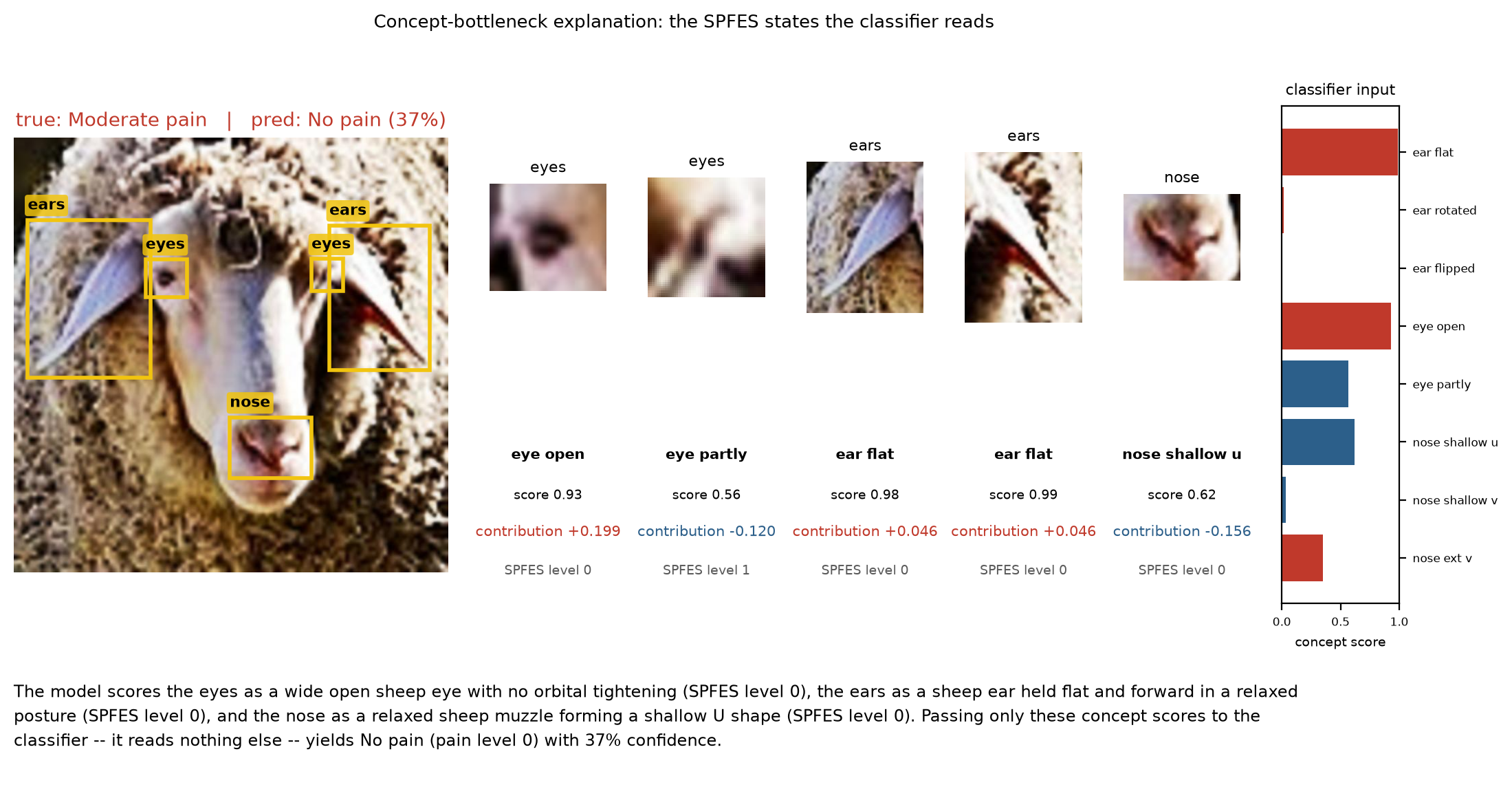}\\[2pt]
{\footnotesize (a)}\\[6pt]
\includegraphics[width=0.62\linewidth]{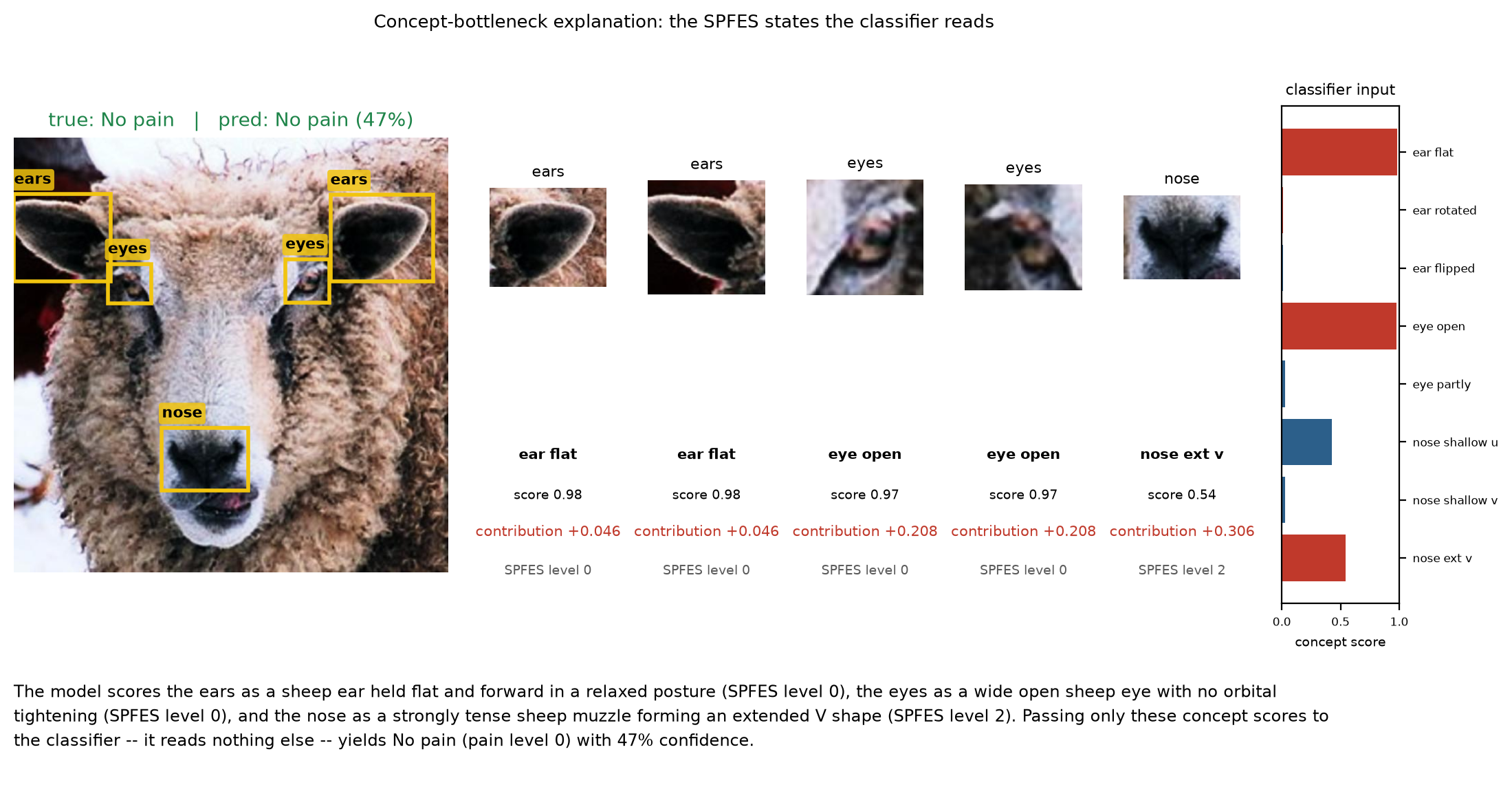}\\[2pt]
{\footnotesize (b)}\\[6pt]
\includegraphics[width=0.62\linewidth]{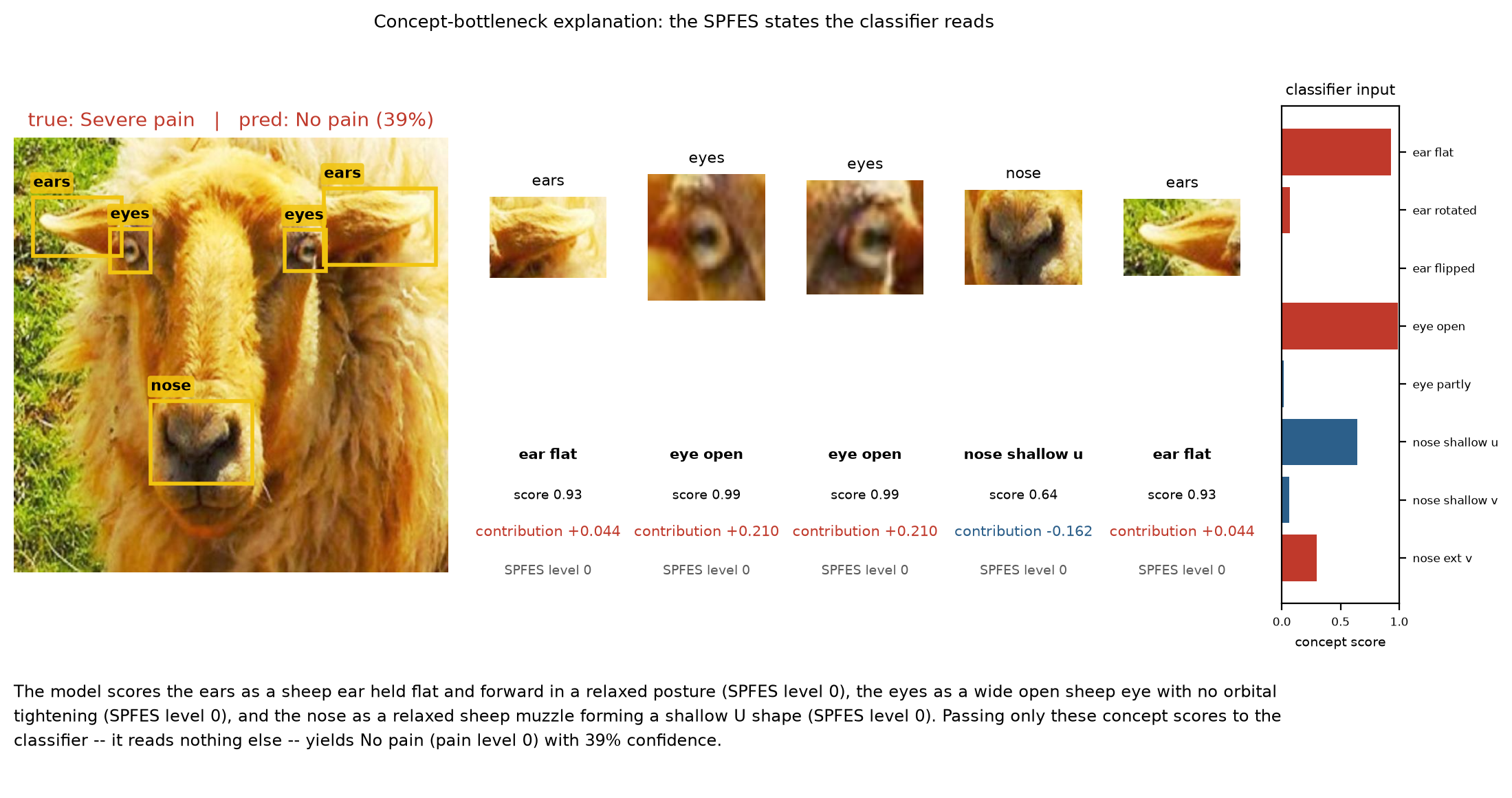}\\[2pt]
{\footnotesize (c)}
\caption{Concept-bottleneck explanations. Each panel shows the detected regions, the SPFES state assigned to each with its score and its contribution to the predicted level, and the concept vector that forms the classifier's
entire input. (a) An animal in moderate pain, predicted pain-free. One eye is correctly scored as partly closed, an SPFES level-1 state, and its contribution of $-0.120$ argues against the no-pain class; it is outweighed by the four
level-0 assignments on the remaining regions. (b) A pain-free animal, correctly predicted, in which the nose is scored extended-V, the most severe nose state, contributing $+0.306$ towards no pain. That state occurs once in the corpus:
the head has learned a detector for a cue that co-occurs with pain-free animals, and the classifier has weighted it accordingly. (c) An animal in
severe pain, predicted pain-free, with every region read as a level-0 state. The explanation is internally consistent with the prediction, and both are wrong. In each panel the error can be traced to specific concept assignments
rather than inferred from a saliency map.}
\label{fig:cbm-qual}
\end{figure*}

Figure~\ref{fig:cbm-qual} shows explanations from the bottleneck. Each panel gives the detected regions, the SPFES state assigned to each with its score, that state's contribution to the predicted level, and the full eight-dimensional concept vector, which is the classifier's entire input. The contributions sum, with the bias, to the predicted logit; nothing is estimated after the fact, and nothing can be named that the classifier did not read. Panel (a) represents the case for which the design is intended, where a moderately painful sheep is correctly identified, and the two cues driving the decision are an ear rotated ($+0.473$) and an eye that is partly closed ($+0.098$). SPFES level-1 states the cues for both cases, specifically indicating what a trained scorer should look for. The nose is read as relaxed and contributes in the opposite direction. A practitioner might verify each claim against the animal. Panel (b) is a correctly identified pain-free sheep in which all three regions are assigned level-0 states, and the explanation is unremarkable in the way it should be. While panel (c) is the failure, and it is the reason we show these panels rather than only aggregate statistics. The sheep is in severe pain and not correctly classified; the nose is scored shallow-U, which is actually the most severe nose state in the scale with a score. Both halves of that sentence are wrong clinically, and they are wrong in a way that cancels.

\paragraph{Concept leakage}
The pattern is systematic rather than anecdotal. Across the twelve test images analyzed, the nose head assigns extended-V to six of the eleven nose regions, every one of which is annotated shallow-U; all six occur on images predicted pain-free, where the assignment contributes between $+0.25$ and $+0.37$ towards the no-pain class. Panel~(b) of Fig.~\ref{fig:cbm-qual} shows one such case. The coordinate is therefore load-bearing and, in a narrow predictive sense, correctly weighted. The head has learned a discriminative visual cue that co-occurs with pain-free animals. The label, however, names a concept it does not detect. This phenomenon is concept leakage \cite{margeloiu2021leakage} in an unusually legible form, because the bottleneck forces the information through a named channel, and the misnaming therefore appears in the explanation rather than remaining hidden in a feature vector. We draw two observations from this phenomenon. First, a concept bottleneck renders a model's reasoning inspectable without rendering it correct. The value of the inspection depends on the concept-level evaluation accompanying it, and a pain-accuracy figure alone would not have surfaced this issue. Second, the behavior is governed by the support available for each concept rather than by the architecture. Severe nose states are intrinsically rare in any welfare corpus collected under field conditions because severely affected animals are uncommon, and models require multiple instances to estimate a concept. Severely affected animals are uncommon, and a concept represented by a single instance cannot be estimated by any model. Concept-level supervision therefore imposes sampling requirements beyond those of image-level labeling, a consideration relevant to the design of future annotation efforts in this domain.

\subsection{Emergent severity ordering}
\label{sub:ordering}

The concept loss \eqref{eq:cbmloss} supervises the state of a region but does not indicate which state is more painful. The severity ordering is therefore free to be recovered from the pain labels alone, and reading it off the classifier weights provides a check on whether the bottleneck has organized itself clinically. Table~\ref{tab:global} provides those weights, centered within the region and averaged over the five folds. For the two regions with adequate support, the recovery is complete. The ear weights are monotone in severity across all three levels. An ear flat (level 0) carries the strongest weight towards no pain, an ear rotated (level 1) towards moderate pain, and an ear flipped (level 2) towards severe pain. The eye weights are likewise correctly oriented, with the eye open favoring no pain and the eye partly closed favoring moderate pain, and they are exactly antisymmetric because the region has only two states whose scores sum to one. Five of the eight orderings are correct, and the three failures are the nose states. The level of this impact is negligible; the ear's contribution to severe pain ranges just $0.20$ in logit units over a class of $24$ samples, but it represents the proper ranking, achieved without severity supervision, in the two areas where data permits assessment. We interpret it as evidence that the bottleneck coordinates act as clinical variables rather than just random eight-dimensional encoding, an observation that the unsupervised ablation described in Section~\ref{sub:cbm-ablation} shows should not be assumed. The limitations of the nose are detailed in Section~\ref{sub:cbm-qual}; the nose head assigns an extended V to the majority of the nose regions it recognizes, despite this state occurring exclusively once in the corpus, and the classifier interpreted that distribution as a sign of absence of pain, providing it the most significant weight in the table ($+0.619$ toward no pain). The weight is rationally justifiable but lacks clinical significance in terms of showing theory leakage when the leaking channel is explicitly defined.

\begin{table*}[!t]
\centering
\caption{Classifier weights $W_c$, centred within region and averaged over the five folds, in logit units. Severity is never supervised; the ear and eye
orderings are nonetheless recovered. Bold marks the concept most strongly weighted towards each pain level; it should be the concept annotated at that
level. Nose weights are fitted on $296$, $35$ and $1$ instances respectively and are discussed in Section~\ref{sub:cbm-qual}.}
\label{tab:global}
\small
\begin{tabular}{llcrrr}
\toprule
Concept & Region & SPFES level & No pain & Moderate & Severe \\
\midrule
ear flat        & ears & 0 & $\mathbf{+0.107}$ & $+0.135$ & $-0.113$ \\
ear rotated     & ears & 1 & $+0.002$ & $\mathbf{+0.568}$ & $+0.022$ \\
ear flipped     & ears & 2 & $-0.109$ & $-0.702$ & $\mathbf{+0.091}$ \\
\midrule
eye open        & eyes & 0 & $\mathbf{+0.248}$ & $-0.190$ & $-0.022$ \\
eye partly closed & eyes & 1 & $-0.248$ & $\mathbf{+0.190}$ & $+0.022$ \\
\midrule
nose shallow-U  & nose & 0 & $-0.257$ & $+0.386$ & $+0.401$ \\
nose shallow-V  & nose & 1 & $-0.362$ & $-0.011$ & $-0.192$ \\
nose extended-V & nose & 2 & $+0.619$ & $-0.375$ & $-0.209$ \\
\bottomrule
\end{tabular}
\end{table*}

\subsection{Ablation of the residual-attention model}
\label{sub:ablation}

Table~\ref{tab:ablation} ablates the residual-attention model under the same learning budget as Table~\ref{tab:sota}. The pattern is unambiguous and unfavorable to the design; every component we added makes agreement worse. Removing the SPFES cross-attention raises $\kappa$ from $0.578$ to $0.595$. Removing the graph raises it to $0.628$, the highest value we obtain from any configuration in this study. Deepening the graph to $K = 3$ lowers it to $0.523$. Removing both the graph and the language leaves the plain region encoder at $0.609$. The individual differences remain within one standard deviation, and we do not claim significance for any single comparison. But the ordering is consistent, it is the same ordering that Table~\ref{tab:sota} produces from independently trained models, and the direction is the opposite of the one the architecture was designed to achieve. Taken with Section~\ref{sub:attrib-results}, which shows the cross-attention has no causal influence on the prediction, the natural reading is that these components add parameters and variance without
adding information.

\begin{table*}[!t]
\centering
\caption{Ablation of the residual-attention model, five-fold cross-validation under the matched $150$-epoch protocol. Every removal improves agreement; deepening the graph degrades it.}
\label{tab:ablation}
\small
\begin{tabular}{lcc}
\toprule
Variant & Cohen's $\kappa$ & Accuracy \\
\midrule
Full model (SPFES x-attn, $K{=}1$)     & $0.578 \pm 0.084$ & $0.814 \pm 0.030$ \\
\quad without SPFES cross-attention    & $0.595 \pm 0.068$ & $0.822 \pm 0.028$ \\
\quad \textbf{without graph} ($K{=}0$) & $\mathbf{0.628 \pm 0.070}$ & $0.839 \pm 0.028$ \\
\quad without graph, max-pool readout  & $0.601 \pm 0.063$ & $0.826 \pm 0.026$ \\
\quad without graph and language       & $0.609 \pm 0.046$ & $0.831 \pm 0.023$ \\
\quad deeper graph ($K{=}3$)           & $0.523 \pm 0.103$ & $0.799 \pm 0.039$ \\
\bottomrule
\end{tabular}
\end{table*}

\subsection{Ablation of the concept bottleneck}
\label{sub:cbm-ablation}

Table~\ref{tab:cbm-ablation} eliminates the bottleneck and contains the most critical experiment in this study. Conceptual supervision is what elevates concepts with importance. To remove the per-region supervision from \eqref{eq:cbmloss}, the architecture, the static text embeddings, and the bottleneck are preserved, resulting in a statistically invariant pain agreement ($0.528 \pm 0.104$ compared to $0.524 \pm 0.077$), whereas concept accuracy reduces from $0.801$ to $0.109$. The probability contained within the regional masks is around $0.36$, indicating that the unsupervised bottleneck is not only uninformative on the SPFES states but also exhibits an anti-correlation with them. The implications should be expressed with precision, and the unsupervised form represents a conceptual limitation in all architectural dimensions. The classifier interprets just eight numerical values, and these values correspond to fixed embeddings of clinical terminology, and an intervention assessment would verify their deterministic importance. It would successfully clear every assessment we conducted in Section~\ref{sub:attrib-results}. However, their coordinates do not align with the states it examines, and its pain agreement fails to provide any evidence of their precision. Architectural conditions ensure that the ideas are employed; only supervision and assessment against concept labels confirm the intended uses.

While the graph compromises accuracy for fidelity, eliminating the message-passing phase shifts the bottleneck in both directions simultaneously. So the concept accuracy increases from $0.801$ to $0.911$, while pain agreement decreases from $0.524$ to $0.446$. Increasing the depth to $K = 3$ yields a concept accuracy of $0.913$ and a kappa value of $0.412$. The mechanism is clear throughout the architecture. Message passing integrates adjacent areas into the representation of each region before the computation of concept scores in \eqref{eq:concept}; therefore, after one iteration, a region's score embodies the entire face rather than just that specific region. This mechanism assists the pain decision, which is a total facial assessment, and adversely affects the conceptual decision, which is by definition localized. An explanation per area derived from a $K = 1$ model is thus somewhat compromised, and the eye score does not just correspond to the eye. This represents a trade-off instead of a flaw, although it is one that conceptual models founded on relational frameworks need to evaluate and report. An expert requiring regional justification should choose for $K = 0$ and agree to $\kappa = 0.446$; conversely, one seeking the most accurate pain estimate the bottleneck can provide should accept marginally imprecise notions at $\kappa = 0.524$. 

\begin{table*}[!t]
\centering
\caption{Ablation of the concept bottleneck, five-fold cross-validation. Removing concept supervision leaves pain agreement untouched while destroying
concept fidelity; removing the graph reverses the trade.}
\label{tab:cbm-ablation}
\small
\begin{tabular}{lccc}
\toprule
Variant & Cohen's $\kappa$ & Accuracy & Concept acc. \\
\midrule
Concept bottleneck ($K{=}1$)   & $0.524 \pm 0.077$ & $0.804 \pm 0.034$ & $0.801 \pm 0.030$ \\
\quad without concept supervision & $0.528 \pm 0.104$ & $0.791 \pm 0.041$ & $\mathbf{0.109 \pm 0.010}$ \\
\quad without graph ($K{=}0$)  & $0.446 \pm 0.091$ & $0.777 \pm 0.038$ & $\mathbf{0.911 \pm 0.015}$ \\
\quad deeper graph ($K{=}3$)   & $0.412 \pm 0.167$ & $0.771 \pm 0.056$ & $0.913 \pm 0.017$ \\
\bottomrule
\end{tabular}
\end{table*}

\subsection{Per-class behavior, representation, and calibration}
\label{sub:perclass}

\begin{figure*}[!t]
\centering
\includegraphics[width=\linewidth]{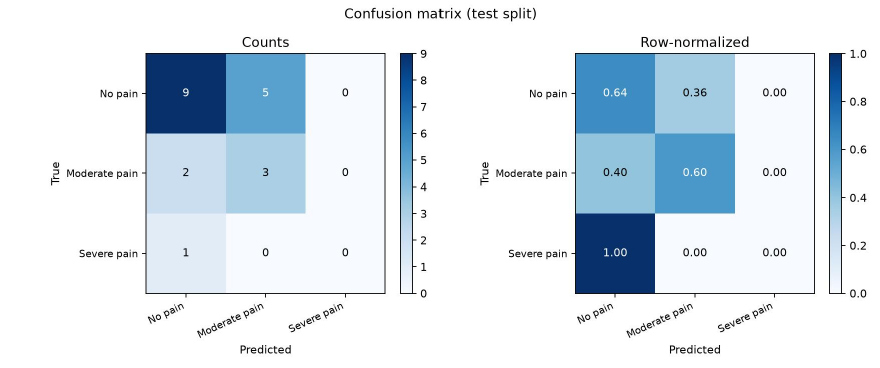}
\caption{Confusion matrix on the held-out partition, as counts (left) and row-normalized (right). Errors are concentrated between adjacent pain levels. The single severe-pain instance is not recovered.}
\label{fig:confusion}
\end{figure*}

\begin{figure}[!t]
\centering
\includegraphics[width=0.9\linewidth]{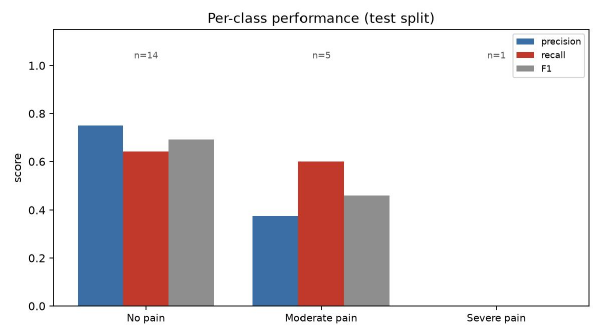}
\caption{Per-class precision, recall, and F1 on the held-out partition, with support ($n$) annotated above each group. The severe-pain class is represented by a single image, and no reliable estimate is possible.}
\label{fig:perclass}
\end{figure}

\begin{figure*}[!t]
\centering
\begin{minipage}{0.48\linewidth}
\centering
\includegraphics[width=\linewidth]{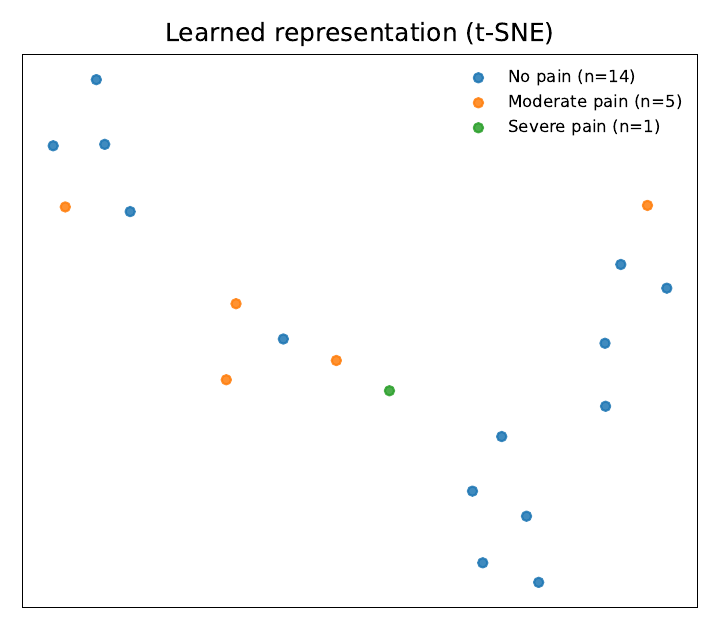}
\end{minipage}\hfill
\begin{minipage}{0.48\linewidth}
\centering
\includegraphics[width=\linewidth]{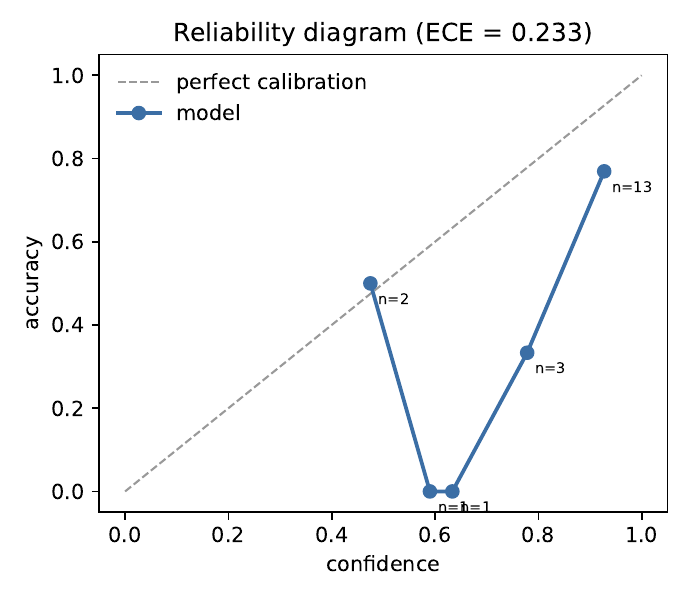}
\end{minipage}
\caption{Left: two-dimensional projection of the graph-level descriptors, colored by true pain level; classes are not cleanly separated. Right:
reliability diagram with expected calibration error $0.233$; bin populations ($n$) are small, so the non-monotonic shape reflects sample size rather than a stable property of the model.}
\label{fig:tsne}
\end{figure*}

Performance exhibits significant asymmetry among classes. Figure~\ref{fig:confusion} illustrates the confusion matrix for a representative fold, while Figure~\ref{fig:perclass} presents the associated per-class scores. The model retrieves no pain at a recall of $0.64$ and moderate discomfort at $0.60$, although it completely fails to identify extreme pain, which is represented by only one instance. The fundamental mistake is in the misinterpretation of adjacent levels, where five images without pain are classified as moderate and two moderate images are classified as no pain, which reflects the anticipated failure pattern for an ordinal scale perceived as unordered classes.  The studied graph-level descriptors are projected into two dimensions in Figure~\ref{fig:tsne}, and the classes do not represent separate clusters. This visualization, with twenty points and one prevalence of severe pain, fails to confirm a conclusion in any direction; it is included for completeness rather than as evidence. Confidence is poorly calibrated (estimated calibration error $0.233$), exhibiting a non-linear reliability profile where predictions based on levels of mid confidence are less precise than those of low confidence. This shape is unusual, and the reasoning behind it is likely due to sample size, where several confidence levels contain one to three predictions. The asymmetry observed per class and the calibration behavior should not be interpreted as properties of the architecture; both are obtained from twenty-four severe-pain images from the small sheep facial expression dataset.

\section{Discussion}
\label{sec:discussion}

\subsection{Relational reasoning does not help on small anatomical graphs}

Under the matched protocol, every graph-based method scored below a graph-free region encoder with mean pooling. The ablation in Table~\ref{tab:ablation} points the same way; removing the graph from our model raises agreement from $0.578$ to $0.628$. Three separate pieces of evidence agree on this. The ablation shows it, the ordering of methods in Table~\ref{tab:sota} shows it, and both two-hop baselines score below the single-hop model. We first trained the ablation with a shorter $40$-epoch budget. All variants then fell within $0.02$ $\kappa$ of one another, which would have supported only the weaker claim that the graph does not help. The matched-budget repetition is what showed that it hurts. The reason appears to be structural. The graph has at most nine nodes covering three region types, and the anatomical adjacency connects them densely. After one propagation step, every node has therefore already collected information from the whole face. Message passing adds little beyond what mean pooling provides, and it adds parameters that small training images cannot constrain. The variance of the results shows the difference clearly. Graph-based methods have fold standard deviations of up to $0.094$, compared with $0.046$ for the graph-free encoder. On this task, extra capacity buys instability rather than better discrimination. 

The scope of this claim needs to be stated carefully, because it is easy to read too much into it. We recognize that graph neural networks are well-suited for facial analysis and that relational structure plays a crucial role in pain assessment. Our claim applies to one particular setting, and that is when the graph is small enough that a single hop reaches every node. The adjacency is fixed by anatomy instead of learned, and the training set contains only a few hundred images. Any of these conditions may not hold elsewhere. A per-landmark design would provide many more nodes and would not saturate after one hop. A dataset 10$\times$ larger would constrain the additional parameters. A learned adjacency might discover couplings that the anatomical prior misses. The claim also does not cover the settings of other works like \cite{11099391}. There, the graph combines per-region pain levels that the detector supplies, instead of inferring pain from appearance. In that role the graph acts as a structured pooling rule over values that already carry the answer, so the present result does not apply to it. What we do claim is that the methodological graph formulations are often applied to facial part graphs and  small structured visual problems in general, without a graph-free baseline trained under the same budget. Our experience finds that such a baseline should be required. It is worth recording how easily we could have missed this opportunity ourselves. A stable result appeared only after the budget was extended and matched across methods, and the graph-free baseline was the last row we added. An evaluation that had stopped earlier would have supported a different and less accurate conclusion.

\subsection{What it takes for an explanation to be worth anything}

Two of our models produce explanations. Only one of them means anything, and the difference was not visible until we ran an intervention test. At first glance, the residual-attention model appears to be the more convincing of the two. It scores higher ($0.578$ against $0.524$). Its attention maps are cleanly organized, its gate stays open, and it produces fluent sentences rather than a fluent sentence that names clinical cues. Suppose we had reported it with the evidence that interpretability papers usually show; a heat map, a gate histogram, and a few qualitative panels. It would have read as a successful language-grounded system, and we would not have known otherwise. The ablation would not have warned us either. Removing the cross-attention results in a cost of only $0.010$ $\kappa$, which, given the size of this corpus, can be easily characterized as redundancy rather than irrelevance. The problem became visible only when we asked a different question. The usual question is how much attention a descriptor receives, and the model answers it readily. The better question is what happens to the prediction when that descriptor is removed. This situation requires an intervention, and the answer is nothing. 

We believe such a test should be routine wherever attention is presented as an explanation. It costs $C$ forward passes. The general point that attention weights need not be explanations is not new \cite{jain2019attention}. What our result adds is a clinical case in which every other indicator pointed the other way: the structure in the attention map, the open gate, and the fluent grounded text. The concept bottleneck earns its explanation in a different way. It makes the clinical vocabulary load-bearing instead of optional. This is a structural guarantee rather than an experimental finding, since the concepts are always relevant when the classifier reads nothing else. The intriguing question therefore moves elsewhere: do the concepts mean what their names say? Section~\ref{sub:cbm-concepts} answers this question with per-concept recovery rather than with a single accuracy figure, and the answer is qualified. Four minority states are detected well above their base rates. We prefer to report that qualified picture rather than the unqualified one the residual model would have allowed. It is also worth noting which model a practitioner should choose. The graph-free encoder is the most accurate and explains nothing. The residual-attention model ranks second but makes statements that are inconsistent with its reasoning. The bottleneck model ranks third and makes accurate statements. For a welfare decision with real consequences, we believe the ordering that matters is the reverse of the accuracy ordering.

\subsection{On reporting concept accuracy}

The imbalance analysis of Section~\ref{sub:cbm-concepts} applies beyond this dataset, and we state it separately because we expect the problem to recur. Clinical annotation is imbalanced almost by definition. Most animals examined are not in pain, so most annotated regions are in their normal state. In our corpus, two states account for $72\%$ of all annotations, and within each region type, the majority state covers between $0.858$ and $0.892$ of the instances. A concept model evaluated by pooled accuracy is therefore graded mainly on how well it reproduces the normal state. That is the one clinical finding for which nobody needs a model. Our model reaches $0.801$ pooled accuracy, which is below the $0.869$ of a constant predictor. On that comparison alone it would look like a failure. Its balanced recall of $0.563$ compared to $0.429$ indicates a $3.5$ to $8.3$ times higher detection rate. So the pain-indicating states describe a model that is doing the useful thing. Reporting pooled concept accuracy without its majority baseline. Therefore, it is uninformative at best, and it can mislead in both directions. It can flatter a model that has learned nothing but the prior. Moreover, it can condemn a model that has given up some prior fitting in exchange for clinically meaningful detection. We instead report per-concept recall against base rates, along with the support for each concept, and summarize it with balanced recall.

\subsection{Efficiency and the path to field deployment (LoRA)}

The LoRA result addresses a practical obstacle, and the welfare monitoring is increasingly planned for mobile and robotic platforms. These include handheld devices, fixed installations at water points \cite{HITELMAN2022106713}, and unmanned aerial vehicles flying over grazing land \cite{10155900,SARWAR2021106219}. Memory and power limitations restrict the deployment of large system models in such systems. They are limited even more in their capacity to be retrained when they meet a new breed, housing system, lighting condition, or camera. Reducing the number of adaptable parameters from $11.5$M to $1.18$M makes adaptation on the device or  the farm much more practical. The cost in agreement stays within one standard deviation. We are explicit about the limits of this argument. Our images are taken from the ground at close range. Facial action units need a spatial resolution that is not available at normal UAV flight altitudes, and we have not tested any part of this system on aerial images. A realistic role for a UAV in this pipeline is to detect and locate animals that deserve closer inspection \cite{SARWAR2021106219}. The facial pain assessment would then be carried out by a sensor at ground level or at low altitude. Before any claim about aerial pain scoring can be made, the minimum resolution at which SPFES cues remain recoverable must be established. We regard this as future work rather than a demonstrated capability.

\section{Limitations}
\label{sec:limitations}

The concept-level limitations are as serious as the image-level ones, and they are sharper. Nose extended-V is clinically the most informative nose state, yet it occurs very few times in maximum annotated regions. The weight in $W_c$ is based on a few examples, which renders it meaningless, and as a result, the nose severity ordering is ineffective. Ear flipped ($n=34$) and nose shallow-V ($n=35$) are recovered above their base rates, but the support is too small for a confident estimate. Only ear rotated ($n=80$) and eye partly closed ($n=104$) rest on enough data for us to treat their recovery as established, and even these are modest. The qualitative panels of Fig.~\ref{fig:cbm-qual} come from the first twelve test images and were not selected for effect. The leak they expose appears in six of the eleven nose regions among those images. Twelve images remain a small window, however, so the panels illustrate the problem rather than establish its extent. 

A second concept-level caveat concerns the bottleneck itself. The per-region scores are a softmax over the states of that region, so the columns of $W_c$ within a region are linearly dependent. The weights are then identified only up to an additive constant per region. We center them before reporting, which removes this constant, but it also means that only comparisons within a region can be interpreted. A statement comparing an ear weight with a nose weight, for example, is not supported by our analysis. Concept bottlenecks that use softmax-normalized groups share this property, and in our reading they do not always account for it. At the image level, the main limitation is the scarcity of severe pain. Twenty-four such images across the corpus provide roughly five per fold, and in some test partitions only a few times. The model fails to recover this class reliably, its per-class metrics remain unstable, and these experiments do not support any conclusions about severe pain. 

The same scarcity limits what the calibration analysis can establish, because the confidence bins contain few samples. The corpus is also small in absolute terms, and differences of a few hundredths in $\kappa$ cannot be resolved, and we have been careful not to claim otherwise. The consistent ordering across the comparison and the ablation supports our qualitative conclusion about relational reasoning, but it does not support fine distinctions between individual methods. The pain scale is ordinal, and we treat it as three unordered classes. The confusion matrix shows that errors concentrate between adjacent levels. An ordinal formulation would penalize such errors appropriately, and Cohen's $\kappa$ already reflects this approach in part. We did not pursue this direction, and we regard it as the most promising untried route to higher accuracy. 

Our evaluation assumes that regional localizations are available. In deployment they would come from a detector whose errors would propagate through the pipeline, and we have not measured how sensitive the system is to detection quality. The detector's own class scheme also encodes a pain level for each region, so a detector-only readout is possible in principle and would make a worthwhile additional baseline. The explanations are faithful to the model's computation, but Section~\ref{sub:attrib-results} shows that an explanation can look well formed while contributing nothing. Their clinical usefulness has not been assessed with practitioners. Whether the justifications improve a stockperson's decisions or their confidence in the system is an empirical question that requires a user study we have not carried out. We regard the lack of a user study as the most important gap between the present work and a validated claim of trustworthiness. Finally, the images are taken from the ground at close range and come from a single annotated corpus. Generalization across breeds, housing systems, lighting conditions, and acquisition platforms remains untested, and the issue applies in particular to aerial imagery.

\section{Conclusion}
\label{sec:conclusion}

We set out to build an interpretable sheep facial pain assessor, grounded in the clinical language of the SPFES. Our model lets each region attend to frozen text embeddings of the clinical descriptors. It produced attention maps with clean anatomical structure. The language gate stayed open. The justifications it generated named specific clinical cues and read fluently. Then we ran an intervention test. Removing an entire descriptor changed the prediction by about $10^{-4}$. The most-attended cue matched the predicted pain level in only a third of regions. None of the diagnostics that interpretability papers usually report would have revealed this. The solution is not to adjust how the language enters the model. It is to remove the appearance bypass altogether. In a concept-bottleneck formulation, the SPFES states are supervised directly by per-region annotations, which image-level pipelines normally discard. The concepts then become causally necessary by construction. This approach costs $0.054$ $\kappa$ against the unfaithful model and $0.085$ against the strongest baseline, which offers no explanation at all. 

The price of interpretability is therefore measured, not argued. In return, the concepts are demonstrably learned. The minority SPFES states, which are the ones that indicate pain, are recovered at $3.5$ to $8.3$ times their base rates. The ear and eye severity orderings emerge on their own, with no supervision of severity. To demonstrate this finding, it was necessary to set aside the metrics by which such models are typically evaluated. Our pooled concept accuracy is 0.801. Consistently naming the majority state of a region results in an accuracy of $0.869$. By that comparison our model looks worse, yet it is the one doing the clinically useful work. Under the imbalance typical of clinical annotation, pooled accuracy mostly measures how well a model reproduces the normal finding. Balanced recall against per-concept base rates does not. We also contribute a protocol-matched, cross-validated benchmark of seven methods on this dataset. Within it, we find that relational reasoning over graphs of at most nine anatomical nodes does not improve pain discrimination over a well-trained region encoder. Taken together, these results suggest that on small clinical corpora the harder problem is not reaching adequate accuracy. It is substantiating any claim about why a model is right. Such claims are easy to make, easy to believe, and cheap to test.


\section*{Data and Code Availability}
The implementation, configuration files, and evaluation scripts are
available from the author on request.


\bibliographystyle{IEEEtran}
\bibliography{references}

@article{MCLENNAN201619,
title = {Development of a facial expression scale using footrot and mastitis as models of pain in sheep},
journal = {Applied Animal Behaviour Science},
volume = {176},
pages = {19-26},
year = {2016},
issn = {0168-1591},
doi = {https://doi.org/10.1016/j.applanim.2016.01.007},
url = {https://www.sciencedirect.com/science/article/pii/S0168159116000101},
author = {Krista M. McLennan and Carlos J.B. Rebelo and Murray J. Corke and Mark A. Holmes and Matthew C. Leach and Fernando Constantino-Casas}
}

@article{NOOR2023100366,
title = {Sheep health behavior analysis in machine learning: A short comprehensive survey},
journal = {Smart Agricultural Technology},
volume = {6},
pages = {100366},
year = {2023},
issn = {2772-3755},
doi = {https://doi.org/10.1016/j.atech.2023.100366},
url = {https://www.sciencedirect.com/science/article/pii/S2772375523001946},
author = {Alam Noor and Murray J. Corke and Eduardo Tovar}
}

@INPROCEEDINGS{11099391,
  author={Noor, Alam and Almeida, Luis and Daoudi, Mohamed and Li, Kai and Tovar, Eduardo},
  booktitle={2025 IEEE 19th International Conference on Automatic Face and Gesture Recognition (FG)}, 
  title={Sheep Facial Pain Assessment Under Weighted Graph Neural Networks}, 
  year={2025},
  volume={},
  number={},
  pages={1-9},
  doi={10.1109/FG61629.2025.11099391}}

@article{NOOR2020105528,
title = {Automated sheep facial expression classification using deep transfer learning},
journal = {Computers and Electronics in Agriculture},
volume = {175},
pages = {105528},
year = {2020},
issn = {0168-1699},
doi = {https://doi.org/10.1016/j.compag.2020.105528},
url = {https://www.sciencedirect.com/science/article/pii/S0168169920306633},
author = {Alam Noor and Yaqin Zhao and Anis Koubaa and Longwen Wu and Rahim Khan and Fakheraldin Y.O. Abdalla}
}

@INPROCEEDINGS{7961768,
  author={Lu, Yiting and Mahmoud, Marwa and Robinson, Peter},
  booktitle={2017 12th IEEE International Conference on Automatic Face \& Gesture Recognition (FG 2017)}, 
  title={Estimating Sheep Pain Level Using Facial Action Unit Detection}, 
  year={2017},
  volume={},
  number={},
  pages={394-399},
  doi={10.1109/FG.2017.56}}

@INPROCEEDINGS{10388128,
  author={Feng, Zejian and Karaskova, Martina and Mahmoud, Marwa},
  booktitle={2023 11th International Conference on Affective Computing and Intelligent Interaction Workshops and Demos (ACIIW)}, 
  title={Open-Sheep-Face: A Comprehensive Application for Sheep Face Analysis and Pain Estimation}, 
  year={2023},
  volume={},
  number={},
  pages={1-3},
  doi={10.1109/ACIIW59127.2023.10388128}}

@INPROCEEDINGS{7477733,
  author={Yang, Heng and Zhang, Renqiao and Robinson, Peter},
  booktitle={2016 IEEE Winter Conference on Applications of Computer Vision (WACV)}, 
  title={Human and sheep facial landmarks localisation by triplet interpolated features}, 
  year={2016},
  volume={},
  number={},
  pages={1-8},
  doi={10.1109/WACV.2016.7477733}}

@article{HITELMAN2022106713,
title = {Biometric identification of sheep via a machine-vision system},
journal = {Computers and Electronics in Agriculture},
volume = {194},
pages = {106713},
year = {2022},
issn = {0168-1699},
doi = {https://doi.org/10.1016/j.compag.2022.106713},
url = {https://www.sciencedirect.com/science/article/pii/S0168169922000308},
author = {Almog Hitelman and Yael Edan and Assaf Godo and Ron Berenstein and Joseph Lepar and Ilan Halachmi}
}

@article{ZHANG2022107452,
title = {Biometric facial identification using attention module optimized YOLOv4 for sheep},
journal = {Computers and Electronics in Agriculture},
volume = {203},
pages = {107452},
year = {2022},
issn = {0168-1699},
doi = {https://doi.org/10.1016/j.compag.2022.107452},
url = {https://www.sciencedirect.com/science/article/pii/S0168169922007608},
author = {Xiwen Zhang and Chuanzhong Xuan and Yanhua Ma and He Su and Mengqin Zhang}
}

@article{LI2023107651,
title = {Combining convolutional and vision transformer structures for sheep face recognition},
journal = {Computers and Electronics in Agriculture},
volume = {205},
pages = {107651},
year = {2023},
issn = {0168-1699},
doi = {https://doi.org/10.1016/j.compag.2023.107651},
url = {https://www.sciencedirect.com/science/article/pii/S016816992300039X},
author = {Xiaopeng Li and Yuyun Xiang and Shuqin Li}
}

@Article{ani13111824,
AUTHOR = {Zhang, Xiwen and Xuan, Chuanzhong and Xue, Jing and Chen, Boyuan and Ma, Yanhua},
TITLE = {LSR-YOLO: A High-Precision, Lightweight Model for Sheep Face Recognition on the Mobile End},
JOURNAL = {Animals},
VOLUME = {13},
YEAR = {2023},
NUMBER = {11},
ARTICLE-NUMBER = {1824},
URL = {https://www.mdpi.com/2076-2615/13/11/1824},
PubMedID = {37889716},
ISSN = {2076-2615},
DOI = {10.3390/ani13111824}
}

@article{YUAN2025101362,
title = {Sheep identification based on face features with the method and practical application analysis},
journal = {Smart Agricultural Technology},
volume = {12},
pages = {101362},
year = {2025},
issn = {2772-3755},
doi = {https://doi.org/10.1016/j.atech.2025.101362},
url = {https://www.sciencedirect.com/science/article/pii/S2772375525005933},
author = {Hongbo Yuan and Zhaohan Liu and Zhenjiang Cai and Yingjie Zhang and Man Cheng}
}

@Article{agriculture14030468,
AUTHOR = {Hao, Min and Sun, Quan and Xuan, Chuanzhong and Zhang, Xiwen and Zhao, Minghui and Song, Shuo},
TITLE = {Lightweight Small-Tailed Han Sheep Facial Recognition Based on Improved SSD Algorithm},
JOURNAL = {Agriculture},
VOLUME = {14},
YEAR = {2024},
NUMBER = {3},
ARTICLE-NUMBER = {468},
URL = {https://www.mdpi.com/2077-0472/14/3/468},
ISSN = {2077-0472},
DOI = {10.3390/agriculture14030468}
}

@INPROCEEDINGS{10155900,
  author={Sarantinoudis, Nikolaos and Arampatzis, George and Valavanis, Kimon P. and Tsourveloudis, Nikos},
  booktitle={2023 International Conference on Unmanned Aircraft Systems (ICUAS)}, 
  title={Unmanned Aerial Vehicles and Livestock Management: An Application in Western Crete}, 
  year={2023},
  volume={},
  number={},
  pages={159-166},
  doi={10.1109/ICUAS57906.2023.10155900}}

@article{SARWAR2021106219,
title = {Detecting sheep in UAV images},
journal = {Computers and Electronics in Agriculture},
volume = {187},
pages = {106219},
year = {2021},
issn = {0168-1699},
doi = {https://doi.org/10.1016/j.compag.2021.106219},
url = {https://www.sciencedirect.com/science/article/pii/S0168169921002362},
author = {Farah Sarwar and Anthony Griffin and Saeed Ur Rehman and Timotius Pasang}
}

@ARTICLE{2016arXiv160902907K,
       author = {{Kipf}, Thomas N. and {Welling}, Max},
        title = "{Semi-Supervised Classification with Graph Convolutional Networks}",
      journal = {arXiv e-prints},
         year = 2016,
        month = sep,
          eid = {arXiv:1609.02907},
        pages = {arXiv:1609.02907},
          doi = {10.48550/arXiv.1609.02907},
archivePrefix = {arXiv},
       eprint = {1609.02907},
 primaryClass = {cs.LG},
       adsurl = {https://ui.adsabs.harvard.edu/abs/2016arXiv160902907K}
}

@inproceedings{
veličković2018graph,
title={Graph Attention Networks},
author={Petar Veličković and Guillem Cucurull and Arantxa Casanova and Adriana Romero and Pietro Liò and Yoshua Bengio},
booktitle={International Conference on Learning Representations},
year={2018},
url={https://openreview.net/forum?id=rJXMpikCZ},
}

@inproceedings{Qi2018LearningHI,
  title={Learning Human-Object Interactions by Graph Parsing Neural Networks},
  author={Siyuan Qi and Wenguan Wang and Baoxiong Jia and Jianbing Shen and Song-Chun Zhu},
  booktitle={European Conference on Computer Vision},
  year={2018},
  url={https://api.semanticscholar.org/CorpusID:51992868}
}

@ARTICLE{11186137,
  author={Yang, Guangrui and Li, Ming and Feng, Han and Zhuang, Xiaosheng},
  journal={IEEE Transactions on Pattern Analysis and Machine Intelligence}, 
  title={Deeper Insights Into Deep Graph Convolutional Networks: Stability and Generalization}, 
  year={2026},
  volume={48},
  number={2},
  pages={1707-1719},
  doi={10.1109/TPAMI.2025.3616350}}

@inproceedings{
oono2020graph,
title={Graph Neural Networks Exponentially Lose Expressive Power for Node Classification},
author={Kenta Oono and Taiji Suzuki},
booktitle={International Conference on Learning Representations},
year={2020},
url={https://openreview.net/forum?id=S1ldO2EFPr}
}

@inproceedings{10.1145/3292500.3330956,
author = {Verma, Saurabh and Zhang, Zhi-Li},
title = {Stability and Generalization of Graph Convolutional Neural Networks},
year = {2019},
isbn = {9781450362016},
publisher = {Association for Computing Machinery},
address = {New York, NY, USA},
url = {https://doi.org/10.1145/3292500.3330956},
doi = {10.1145/3292500.3330956},
booktitle = {Proceedings of the 25th ACM SIGKDD International Conference on Knowledge Discovery \& Data Mining},
pages = {1539–1548},
numpages = {10},
location = {Anchorage, AK, USA},
series = {KDD '19}
}

@ARTICLE{2021arXiv210206966B,
       author = {{Baranwal}, Aseem and {Fountoulakis}, Kimon and {Jagannath}, Aukosh},
        title = "{Graph Convolution for Semi-Supervised Classification: Improved Linear Separability and Out-of-Distribution Generalization}",
      journal = {arXiv e-prints},
         year = 2021,
        month = feb,
          eid = {arXiv:2102.06966},
        pages = {arXiv:2102.06966},
          doi = {10.48550/arXiv.2102.06966},
archivePrefix = {arXiv},
       eprint = {2102.06966},
 primaryClass = {cs.LG},
       adsurl = {https://ui.adsabs.harvard.edu/abs/2021arXiv210206966B}
}

@InProceedings{pmlr-v139-radford21a,
  title = 	 {Learning Transferable Visual Models From Natural Language Supervision},
  author =       {Radford, Alec and Kim, Jong Wook and Hallacy, Chris and Ramesh, Aditya and Goh, Gabriel and Agarwal, Sandhini and Sastry, Girish and Askell, Amanda and Mishkin, Pamela and Clark, Jack and Krueger, Gretchen and Sutskever, Ilya},
  booktitle = 	 {Proceedings of the 38th International Conference on Machine Learning},
  pages = 	 {8748--8763},
  year = 	 {2021},
  editor = 	 {Meila, Marina and Zhang, Tong},
  volume = 	 {139},
  series = 	 {Proceedings of Machine Learning Research},
  month = 	 {18--24 Jul},
  publisher =    {PMLR},
  url = 	 {https://proceedings.mlr.press/v139/radford21a.html}
}

@inproceedings{koh2020cbm,
  author    = {Koh, Pang Wei and Nguyen, Thao and Tang, Yew Siang and
               Mussmann, Stephen and Pierson, Emma and Kim, Been and
               Liang, Percy},
  title     = {Concept Bottleneck Models},
  booktitle = {Proc. Int. Conf. Machine Learning (ICML)},
  year      = {2020}
}

@inproceedings{jain2019attention,
  author    = {Jain, Sarthak and Wallace, Byron C.},
  title     = {Attention is not Explanation},
  booktitle = {Proc. Conf. North American Chapter of the Association for
               Computational Linguistics (NAACL)},
  year      = {2019}
}

@inproceedings{margeloiu2021leakage,
  author    = {Margeloiu, Andrei and Ashman, Matthew and Bhatt, Umang and
               Chen, Yanzhi and Jamnik, Mateja and Weller, Adrian},
  title     = {Do Concept Bottleneck Models Learn as Intended?},
  booktitle = {ICLR Workshop on Responsible AI},
  year      = {2021}
}

\end{document}